\documentclass[acmsmall]{acmart}

\usepackage{caption}
\usepackage{subcaption}
\usepackage{amsmath}

\AtBeginDocument{%
  \providecommand\BibTeX{{%
    \normalfont B\kern-0.5em{\scshape i\kern-0.25em b}\kern-0.8em\TeX}}}

\setcopyright{acmlicensed}
\acmJournal{TIOT}
\acmYear{2019} \acmVolume{1} \acmNumber{1} \acmArticle{1} \acmMonth{1} \acmPrice{15.00}

\begin{document}

\title{Bonseyes AI Pipeline - bringing AI to you} 
\subtitle{\textbf{End-to-end integration of data, algorithms and deployment tools}}


 \author{Miguel de Prado}
 \affiliation{
   \institution{Haute Ecole Arc Ingenierie; HES-SO $/$} 
   \institution{Integrated Systems Lab, ETH Zurich}
   \country{Switzerland}}
 \email{miguel.deprado@he-arc.ch}
 
 \author{Jing Su} 
 \affiliation{%
  \institution{School of Computer Science \& Statistics,}
  \institution{Trinity College Dublin}
  \country{Ireland}}
 \email{jing.su@tcd.ie}

 \author{Rabia Saeed}
 \affiliation{%
   \institution{Haute Ecole Arc Ingenierie; HES-SO}
   \country{Switzerland}
 }
 \email{rabia.saeed@he-arc.ch}

 \author{Lorenzo Keller} 
 \affiliation{%
  \institution{Nviso}
  \country{Switzerland}}
 \email{lorenzo.keller@nviso.ai}

 \author{Noelia Vallez} 
 \affiliation{%
  \institution{Universidad de Castilla - La Mancha}
  \country{Spain}}
 \email{Noelia.Vallez@uclm.es}
 
  \author{Andrew Anderson} 
 \affiliation{%
  \institution{School of Computer Science \& Statistics,}
  \institution{Trinity College Dublin}
  \country{Ireland}}
 \email{andersan@cs.tcd.ie}
 
  \author{David Gregg} 
 \affiliation{%
  \institution{School of Computer Science \& Statistics,}
  \institution{Trinity College Dublin}
  \country{Ireland}}
 \email{david.gregg@cs.tcd.ie}

  \author{Luca Benini}
 \affiliation{
   \institution{Integrated Systems Lab, ETH Zurich}
   \country{Switzerland}}
 \email{lbenini@iis.ee.ethz.ch}
 
   \author{Tim Llewellynn} 
 \affiliation{%
  \institution{Nviso}
  \country{Switzerland}}
 \email{tim.llewellynn@nviso.ch}
 
   \author{Nabil Ouerhani}
 \affiliation{%
   \institution{Haute Ecole Arc Ingenierie; HES-SO}
   \country{Switzerland}}
 \email{nabil.ouerhani@he-arc.ch}
 
   \author{Rozenn Dahyot} 
 \affiliation{%
 \institution{School of Computer Science \& Statistics,}
  \institution{Trinity College Dublin}
  \country{Ireland}}
 \email{Rozenn.Dahyot@tcd.ie}
\renewcommand\shortauthors{First Author. et al}
 
  \author{Nuria Pazos}
 \affiliation{%
   \institution{Haute Ecole Arc Ingenierie; HES-SO}
   \country{Switzerland}}
 \email{nuria.pazos@he-arc.ch}

\renewcommand{\shortauthors}{de Prado, Su, Saeed, et al.}

\begin{abstract}
Next generation of embedded Information and Communication Technology (ICT) systems are interconnected and collaborative systems able to perform autonomous tasks. The remarkable expansion of the embedded ICT market, together with the rise and breakthroughs of Artificial Intelligence (AI), have put the focus on the \textit{Edge} as it stands as one of the keys for the next technological revolution: the seamless integration of AI in our daily life. However, training and deployment of custom AI solutions on embedded devices require a fine-grained integration of data, algorithms, and tools to achieve high accuracy and overcome functional and non-functional requirements. Such integration requires a high level of expertise that becomes a real bottleneck for small and medium enterprises wanting to deploy AI solutions on the \textit{Edge} which, ultimately, slows down the adoption of AI on applications in our daily life.

In this work, we present a modular AI pipeline as an integrating framework to bring data, algorithms, and deployment tools together. By removing the integration barriers and lowering the required expertise, we can interconnect the different stages of particular tools and provide a modular end-to-end development of AI products for embedded devices. Our AI pipeline consists of four modular main steps: \textit{i)} data ingestion, \textit{ii)} model training, \textit{iii)} deployment optimization and, \textit{iv)} the IoT hub integration. To show the effectiveness of our pipeline, we provide examples of different AI applications during each of the steps. Besides, we integrate our deployment framework, Low-Power Deep Neural Network (LPDNN), into the AI pipeline and present its lightweight architecture and deployment capabilities for embedded devices. Finally, we demonstrate the results of the AI pipeline by showing the deployment of several AI applications such as keyword spotting, image classification and object detection on a set of well-known embedded platforms, where LPDNN consistently outperforms all other popular deployment frameworks.
\end{abstract}



\maketitle

\section{Introduction}
Embedded Information and Communication Technology (ICT) systems are experiencing a technological revolution \cite{ICT}. Embedded ICT systems are interconnected and collaborative systems able to perform smart and autonomous tasks and will soon pervade our daily life. The rapid spread of the embedded ICT market and the remarkable breakthroughs in Artificial Intelligence (AI) are leading to a new form of distributed computing systems where edge devices stand as one of the keys for the spread of AI in our daily life \cite{shafique2018overview}.

The rapid adoption of Deep Learning techniques has prompted the emergence of several frameworks for the training and deployment of neural networks. Frameworks such as Caffe \cite{Caffe}, TensorFlow \cite{TF} and PyTorch \cite{Pytorch} have become the most popular training environments by providing great flexibility and low complexity to design and train neural networks. These frameworks also support the deployment of the trained networks, though such deployment is cloud-oriented which makes it impractical for resource-constrained devices operating on the \textit{Edge}. Hence, a second generation of frameworks has come forth to cover such constrained requirements, e.g., TF Lite \cite{TFlite}, ArmCL \cite{armcl}, NCNN \cite{ncnn}, TensoRT \cite{tensorrt} and, Core ML \cite{CoreML}. Such deployment frameworks offer inference engines with meaningful compression and run-time optimizations to reduce the computational and memory requirements that neural networks demand and thus, meet the specifications for mobile and embedded applications.

\begin{figure}[h]
    \includegraphics[width=0.6\textwidth]{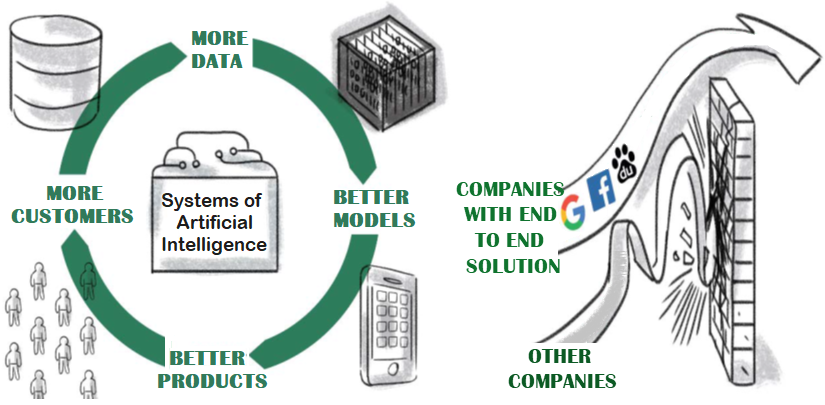}
    \caption{\textbf{AI pipeline and data wall.}
    Companies with end to end solutions outperform the ones unable to follow the complete cyclic process.}
    \label{fig:data_wall}
    \vspace{-0.3cm}
\end{figure}

Developing embedded AI solutions for Computer Vision \cite{huval2015empirical} or Speech Recognition \cite{graves2013speech} implies a significant engineering and organizational effort that requires large investments and development. Although the above-mentioned Deep Learning frameworks are powerful tools, a successful development of custom AI solutions on the Edge requires a fine-grained integration of data, algorithms, and tools to train and deploy neural networks efficiently. Such integration requires a high level of expertise to overcome both software and hardware requirements as well as the fragmentation of AI tools \cite{fragmentation}, e.g., massive and distributed computation for training while reduced and constraint resources for inference, hardship in portability between frameworks, etc. All previous examples represent a real bottleneck for small and medium enterprises who would like to deploy AI solutions on the resource-constrained devices. Only large companies, e.g., Google \cite{google} and Apple \cite{apple}, can build end-to-end systems
for the optimization of Deep Learning applications and are taking the lead of the AI industry \cite{lead}, see Fig. \ref{fig:data_wall}.

As an alternative to monolithic and closed solutions, we form part of the Bonseyes project \cite{Bonseyes}, a European collaboration to facilitate Deep Learning to any stakeholder and reduce development time. We propose an AI pipeline as a way to overcome the aforementioned technological barriers where data, algorithms, and deployment tools are brought together to produce an end-to-end system. We focus on the development of AI solutions for embedded devices, e.g., CPU Arm Cortex-A, embedded GPU, or DSPs, that feature sufficient computational power to run such solutions but need to be efficiently tailored to employ them in real-time applications. Our proposed AI pipeline provides key benefits such as the reusability of custom and commercial tools thanks to its dockerized API, and the flexibility to add new steps to the workflow. 
Thus, the contributions of our work are the following:
\begin{itemize}
    \item We present a modular AI pipeline as an integrating framework to bring data, algorithms, and deployment tools together. By removing the integration barriers and lowering the required expertise, we open up an opportunity for many stakeholders to take up AI custom solutions for embedded devices.
    \item Our AI pipeline encourages the reusability of particular tools by interconnecting them on different stages to provide a modular end-to-end development. The AI pipeline consists of four main steps: \textit{i)} data ingestion, \textit{ii)} model training, \textit{iii)} deployment optimization and, \textit{iv)} IoT hub integration, which we illustrate giving different examples of AI applications.
    \item We integrate deployment framework (LPDNN) into the AI pipeline and present its lightweight architecture and deployment capabilities for embedded devices. Further, we show the deployment of several AI applications such as keyword spotting, image classification and object detection on a set of embedded heterogeneous platforms. Finally, we evaluate LPDNN against a range of popular deployment frameworks where LPDNN consistently outperforms all other frameworks.
\end{itemize}

The paper is organized as follows: in Section 2, we detail the State-of-the-Art. In Section 3, the Bonseyes AI pipeline architecture is introduced. Section 4 describes the data ingestion process. Section 5 presents the training of neural networks. In section 6, the deployment optimization is detailed. Sections 7 introduces IoT hub integration and finally, in section 8 and 9, we present the results and conclusions.

\section{Related work}
In this section, we introduce the state-of-the-art of AI pipelines and frameworks. We take a bottom-up approach starting from those works that propose single services, e.g., training or deployment, to then step up to those works presenting end-to-end solutions.

\subsection{Deep Learning Platforms for High-Performance Computing (HPC)} \label{sec:soa_cloud}
Open-source Deep Learning frameworks have developed fast in recent years. These frameworks bring much convenience to AI project development as they provide the tools to design and train neural networks with available libraries and built-in optimizers, instead of coding from scratch. Here, we present a list of popular frameworks oriented for HPC-based AI applications.

\begin{itemize}
\item Google's TensorFlow (TF) is a powerful framework that provides APIs in multiple languages. TF is built for numerical computation using dataflow graphs in which nodes represent mathematical operations, and graph edges represent multi-dimensional data arrays \cite{TF}.
\item Caffe was developed by Berkeley AI Research (BAIR) and by community contributors\cite{Caffe}. As a pure C++ library, Caffe features expressive architecture, modularity design, and speed of training and inference. Command line, Python, and Matlab interfaces are provided.
\item PyTorch is a scientific framework based on Torch, a C-based library with its scripting language LuaJIT \cite{Torch}. PyTorch can be used as a native library and use popular Python libraries \cite{Pytorch}. PyTorch employs dynamic computation graphs, which offers great flexibility.
\item Microsoft Cognitive Toolkit (CNTK) is an open-source toolkit for commercial-grade distributed deep learning \cite{CNTK}. CNTK can be used as a library in Python, C\#, or C++ programs.
\item Apache MXNet is an open-source framework suited for flexible research prototyping and production \cite{MXNet}. It features hybrid front-end, distributed training, and rich language bindings.
\item Keras is a high-level neural networks API, written in Python and capable of running on top of TensorFlow, CNTK, or Theano \cite{Keras}. Keras is designed for easy and fast prototyping.
\item ONNX is an open ecosystem that provides an open source format for AI models \cite{onnx}\cite{onnx_1}. ONNX greatly improves interoperability between different deep learning frameworks and is becoming a standard for model exchange.
\end{itemize}

Our AI pipeline currently supports Caffe and PyTorch off-the-shelf and is compatible with any of the other frameworks by packaging and integrating them in the second step of the pipeline. Thus, we provide flexibility for the user to select the most suitable training framework.

\subsection{Edge-oriented Deep Learning Frameworks}
\label{sec:soa_edge}
HPC-oriented Deep Learning frameworks facilitate  neural network training and deployment. However, they are not efficient for deployment on resource-constrained devices. To address the need of deployment optimization, a new set of frameworks has appeared to boost the run-time performance of trained models.
\begin{itemize}
\item TensorFlow Lite is a framework tailored for on-device inference \cite{TFlite}. TF Lite does not support the training of neural network directly, users need to convert a trained TF model into the TF Lite format. Thereafter, on-device inference can be carried out through TF Lite Interpreter.
\item Caffe2 is a lightweight deep learning framework focusing the deployment of AI on mobile devices \cite{caffe2fb}. Built on the original Caffe, it comes with C++ and Python API's providing speed and portability. Caffe2 is now a part of PyTorch \cite{caffe2}.
\item Android Neural Networks API (NNAPI) is a C API designed for deployment on Android devices \cite{NNAPI}. NNAPI works as a base layer of functionality which is directly used by an Android app. NNAPI supports and optimizes pre-trained models from TF or Caffe2.
\item Arm Compute Library (CL) is a framework that collects low-level functions optimized for Arm CPUs and GPUs \cite{armcl}. This library is built specially to accelerate image processing, computer vision, and machine learning tasks. 
\item Intel OpenVINO toolkit is a comprehensive toolkit for quickly developing computer vision applications \cite{OpenVINO}. It is designed to maximizing CNNs performance on Intel hardware.
\item Tencent NCNN is a high-performance neural network inference framework optimized for mobile platforms \cite{ncnn}. Currently, a selection of Tencent apps integrate NCNN and making it a popular tool for phone app developers. Most commonly used CNN networks are supported.
\item Alibaba Mobile Neural Network (MNN) is a lightweight inference engine \cite{mnn}. MNN is built to accelerate inference on mobile and embedded devices on iOS and Android. Tensorflow, Caffe and ONNX architectures are supported.
\item Nvidia TensorRT is built on CUDA and it provides capabilities to optimize a trained model for data centre, embedded devices or autonomous driving platforms \cite{tensorrt}. TensorRT is widely compatible with neural network models trained in major frameworks.
\item Tengine by Open AI Lab is a lite, high-performance, and modular inference engine for embedded devices \cite{tengine}. Most Convolutional network operators are supported. Caffe, ONNX, Tensorflow and MXNet models can be loaded directly by Tengine.
\item Core ML is Apple's tool to create or convert machine learning models for iOS apps \cite{CoreML}. Core ML APIs enable on-device prediction with user data as well as on-device training. Besides, it supports models from Caffe, Keras and conversions from TensorFlow and MXNet \cite{CoreML_conv}. 
\end{itemize}
Our Bonseyes AI pipeline relies on LPDNN for the deployment on emdebbed devices. LPDNN provides optimized, portable, and light implementations for AI solutions across heterogeneous platforms. Besides, it can integrate third-party libraries or inference engines into its architecture, which makes it very flexible for custom platforms.

\subsection{End-to-end AI Pipelines}

In Section \ref{sec:soa_cloud} and \ref{sec:soa_edge} we have reviewed the frameworks for HPC- and edge-oriented AI-applications. However, these frameworks are oriented for developers and not as off-the-shelf products for end users. In this section we introduce the advances of \textit{AI as a Service} (or \textit{AI pipeline}).
\par

The \textit{leaders} in providing AI services are Google, Amazon, and Microsoft. Google AI Platform \cite{google_ai} and AI Tools \cite{google_ai_tools} offer great convenience to build AI pipeline with TensorFlow as a backend. These platforms cover every step from data ingestion to model deployment, and they empower customer's AI applications for production. Amazon SageMaker \cite{sagemaker} \cite{amazon_ml} is an end-to-end AI pipeline that features easy deployment of machine learning models on AWS at any scale. 
Microsoft Azure \cite{azure_ml} is an enterprise-grade service designed to accelerate machine learning cycle. Azure highlights a machine learning interpretability toolkit to explain model outputs during training and inference phases \cite{azure_interpret}. Moreover, Azure has wide compatibility with open-source frameworks. Overall, these AI pipelines provide a fast solution to build up an AI application out of training models, but their deployment is mainly based on the cloud.


Next, we look into AI pipelines that provide edge-oriented deployment. Microsoft Azure IoT Edge \cite{azure_iot} enables direct deployment of business logic on edge devices via Docker containers. However, this service is clearly in public preview as many tools are under development and limited to specific options.
Amazon AWS IoT Greengrass \cite{Greengrass} provides local inference on edge devices while depending on its cloud for management. They provide support for a range of edge devices but the service is only available in few regions \cite{GreengrassRegion} and under amazon format. 

Google AI Platform \cite{google_ai} also aims to edge deployment via TF Lite which makes the pipeline very competitive and complete. On the other hand, the pipeline lacks flexibility to use or integrate external tool and constrains the user to employ theirs. This fact bring incompatibility issues with user custom systems and drops in performance as we explain later in Section \ref{comparison-embedded-result}. 

\par


Apart from enterprise-level general purpose AI services, we also look into an edge-oriented AI service for vision tasks. Eugene is created as a suite of machine intelligence services toward IoT applications \cite{Eugene}. With components of data labeling, model training, deployment optimization, and IoT integration, Eugene is an excellent example of AI service. The authors show how to tailor deep neural networks to gain efficiency and introduce a scheduling algorithm to select the best network depth at run time. 
However, Eugene's IoT service pipeline is limited in a few aspects. First, the scheduler is designed to tune the depth of ResNet, which raises questions about how effective this design works for other network architectures. Network reduction methods such as pruning may work in Eugene's pipeline but with significant refactoring. Besides, very little information is available on the supported hardware platforms and the deployment inference engine. Neither did the authors mention system integration or support issues, which can be crucial for the successful deployment of AI service. Bonseyes comes in to address all these drawbacks. A detailed explanation of our proposed AI pipeline is provided in the following sections.


\section{BONSEYES AI PIPELINE ARCHITECTURE} \label{pipe archi}
We introduce the fragmentation and obsolescence of tools and propose the pipeline as s solution to create end-to-end environments for deep learning solutions.
\subsection{Fragmentation and obsolescence of tools}
The rapid evolution of technology in the field of machine learning requires companies to update their environments continuously. Being able to upgrade to the latest algorithm or technology is critical to maintaining the leadership in their field. Besides, upgrading existing data processing tools is a very demanding task. Differences in library versions, run-time environments, and formats need to be handled. These challenges can consume a significant amount of time and introduce hard-to-solve bugs. Moreover, companies are often interested in acquiring the off-the-shelf code or complete pipelines to reduce their costs and to acquire advanced technology that they would not be able to develop in-house. Integrating externally generated code is an even more challenging task as it is very likely that what is acquired is not compatible with the existing pipeline. \par

\subsection{Pipeline as a Solution}
One of the main objectives of the Bonseyes AI pipeline framework is to alleviate this problem and link fragmented environments by defining a way to split the AI pipeline in reusable components, define interfaces for interoperability and provide a documented reference implementation of the interfaces to accelerate development. The main goals of the pipeline are twofold; on one side, to isolate the different parts of the pipeline along with their dependencies, and on the other hand, to insert in the glue code that combines them explicitly. The Bonseyes AI pipeline framework is structured around three concepts:
\begin{itemize}
    \item \textbf{Tool:} A software component that performs a specific function in the pipeline. An example of a tool is a software that is capable of training a model from a training dataset.
    \item \textbf{Artifact:} The product of the execution of a tool, e.g. models, datasets. It can be an output of the pipeline or an intermediate result that is processed by other tools.
    \item \textbf{Workflow:} A declarative pipeline description that lists the tools that need to be used and the artifacts that need to be created. For example, a workflow may import a dataset and use it to train a model.
\end{itemize}

\begin{figure}
    \includegraphics[width=0.6\textwidth]{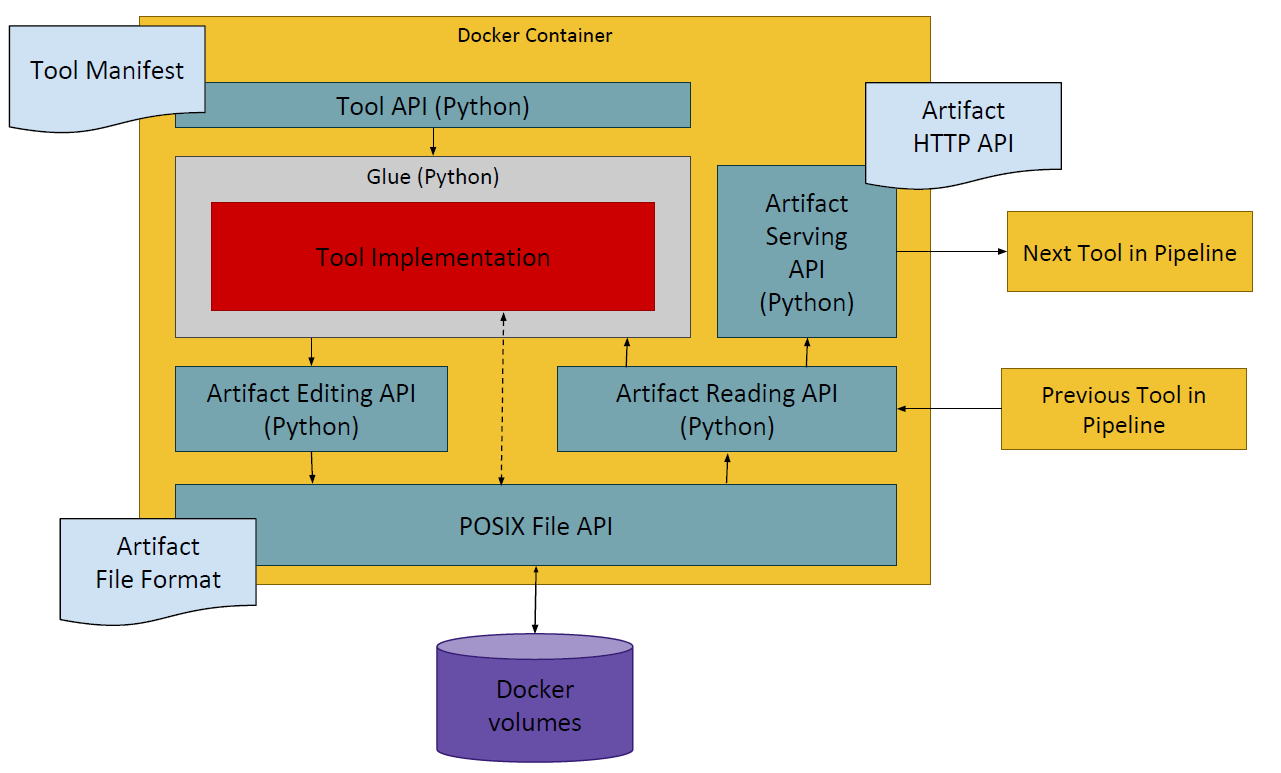}
    \caption{\textbf{Docker container} including all submodules and interfaces.}
    \label{fig:container}
    \vspace{-0.2cm}
\end{figure}

The AI pipeline relies on Docker containers \cite{docker} to package all software dependencies needed to run a tool, see Fig. \ref{fig:container}. Besides, a high-level HTTP API is defined to control the workflows and tools. The AI pipeline provides a collection of standard formats that define on-disk serialization and HTTP REST API for them. In addition, the user can define additional formats with different interfaces to adapt to new designs. 

\subsection{End-to-end AI Pipeline}
\begin{figure}
    \includegraphics[width=0.9\textwidth]{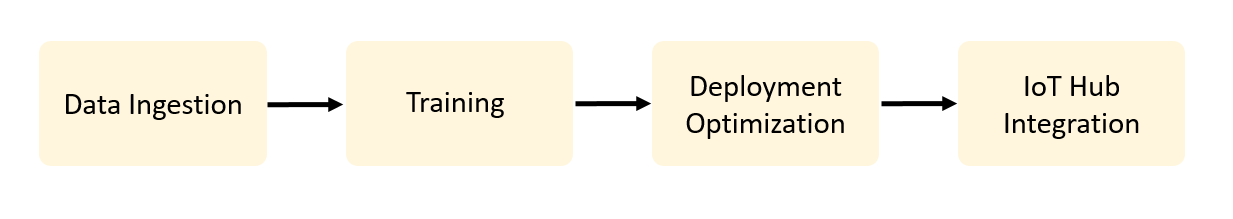}
    \caption{\textbf{Bonseyes AI Pipeline}: Data ingestion, Training, Deployment Optimization and IoT Hub Integration.}
    \label{fig:workflow}
    \vspace{-0.2cm}
\end{figure}

Based on the previous concepts, we propose an end-to-end AI pipeline to develop and deploy Deep Neural Network solutions on embedded devices. We propose a modular AI pipeline architecture which contains four modular main tasks: \textit{i)} Data ingestion, \textit{ii)} Model training, \textit{iii)} Deployment on constraint environment and, \textit{iv)} IoT hub integration, see Fig. \ref{fig:workflow}. The four main tasks may be decomposed into several steps that can be later accomplished by a set of tools. The number of tools for each step is variable since it is possible to have several of them with the same purpose but using different frameworks or dealing with data from various AI challenges such as image classification, object detection or KWS. Moreover, some subtasks may also be optional. For example, Data Partitioning may not be required if the raw data was already partitioned. 

Fragmentation quickly occurs between these tasks as each one needs and employs specific formats to operate. Therein, artifacts come into play as they represent the way by which data can be stored and exchanged between tools. Artifacts must follow a definition to standardize them for each problem type. Hence, tools define their inputs and outputs according to these artifact definitions. This fact makes tools with the same input and output definitions to be easily interchangeable and leads to modularized and reusable pipelines. 

The end-to-end pipeline can be executed following a workflow definition from collecting the data until IoT hub integration. A workflow specifies the steps that are required to obtain the final result. This step involves describing which tools are used and in which order, and how the output artifacts of one tool are used to feed another one. 

In the Bonseyes framework, four different DNN-based AI applications and their corresponding artifact definitions are currently available for the most common AI challenges, although it is possible to add more if needed. These are: image classification, face recognition, keyword spotting and, object detection. In the following sections, we introduce and explain the four main steps of the Bonseyes AI pipeline showing different AI applications on each step.

\begin{figure}
    \includegraphics[width=0.9\textwidth]{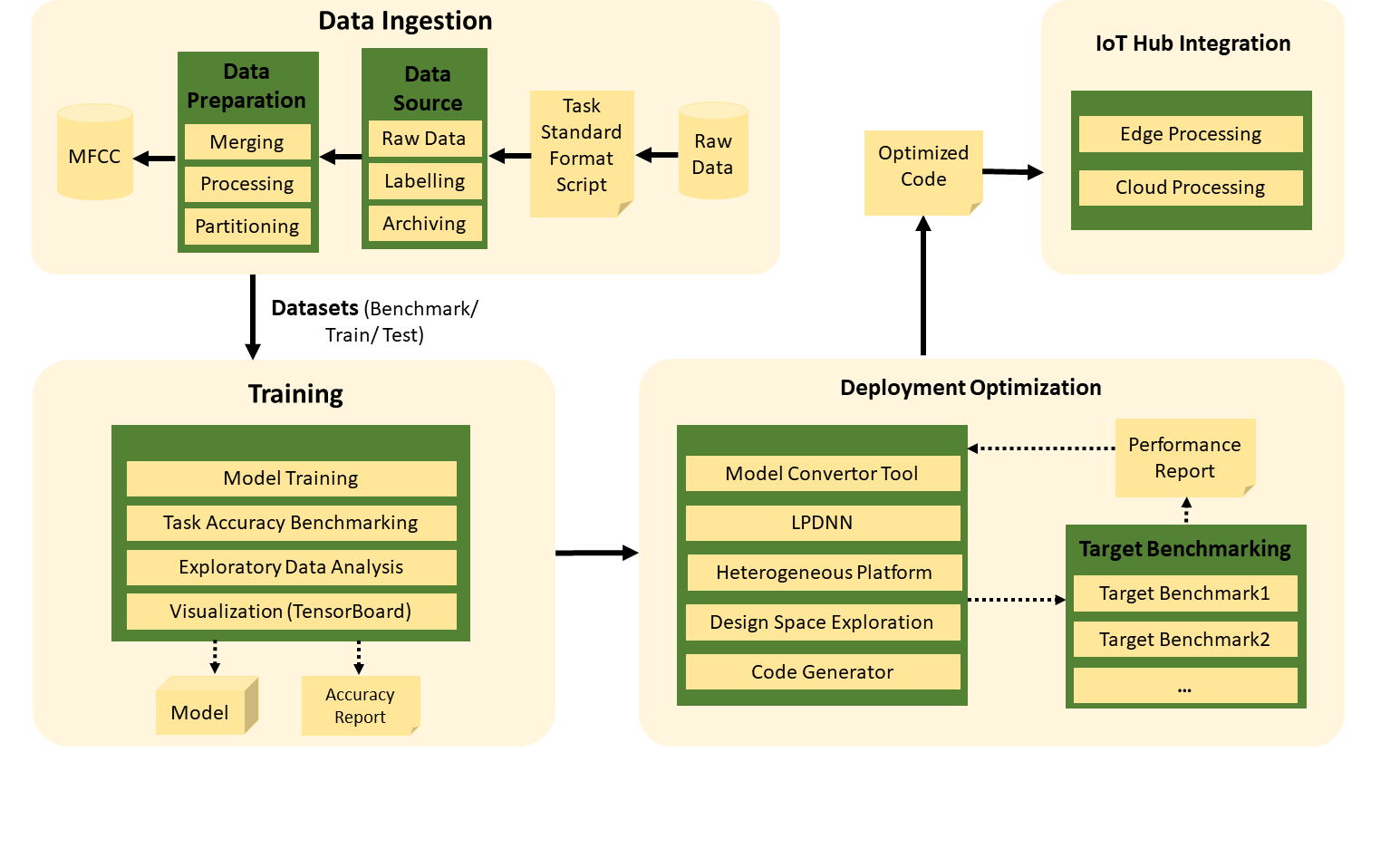}
    \caption{\textbf{Bonseyes AI Pipeline Architecture and functional units.}}
    \label{fig:kws_pipeline}
\end{figure}

\section{AI DATA INGESTION (1/4)}
Data Ingestion is the first step in the AI pipeline, see Fig. \ref{fig:kws_pipeline}. It entails the complete process from acquiring the raw data to have it prepared for model training. The first step of this process involves parsing the raw data and representing it in a standardized format to be used in the next steps according to the problem definition, i.e., classification, KWS, etc. Thus, audio or images are stored together with their annotations. This step is crucial since data is an essential part of the AI pipeline and is shared across multiple stages. However, standardizing it is a tedious task due to the variety of file formats and the lack of a consistent annotation representation. This is the reason why the Bonseyes AI pipeline API provides a set of mechanisms to reduce the efforts when importing new data. Collecting it from the internet or a local hard drive is already supported by specifying only where the resource is located. Moreover,  API also manages how the resulting dataset artifact is stored.

After importing the data into the pipeline, a processing step may be carried out to prepare it for model training. This process includes operations such as image resizing, normalization or face alignment among others. This process can be accomplished by more than one step, which may vary according to the data, the model to be trained, or even the application. Finally, the dataset may also require to be partitioned into training, validation, and test sets depending on the needs. This partitioning is not necessary if it was already done in the raw data, but a large number of public datasets are stored in a single compressed file that requires further processing.

\begin{figure}
    \includegraphics[width=\textwidth]{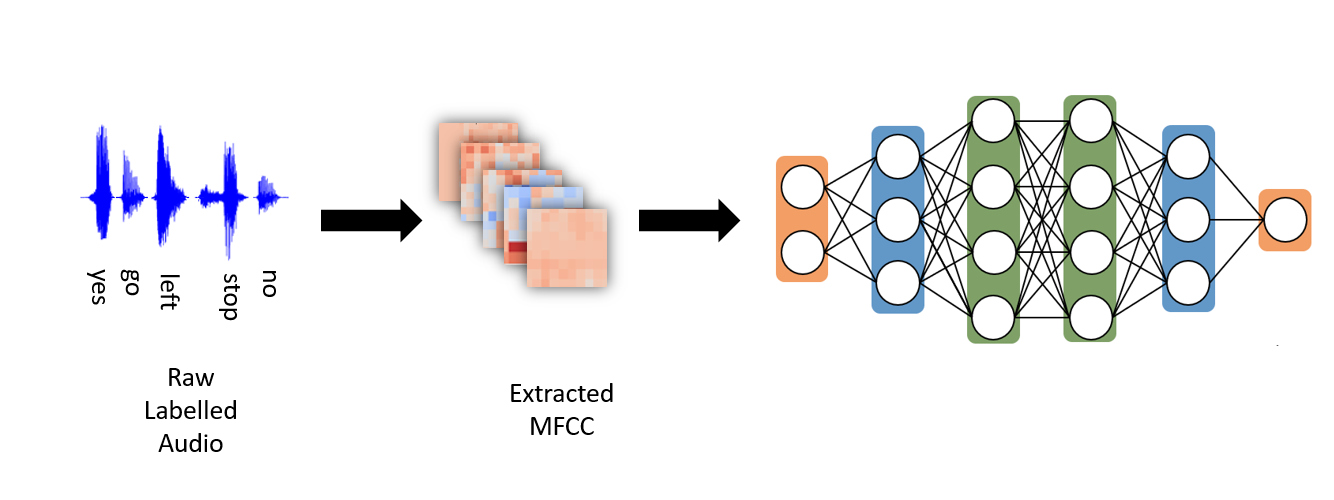}
    \caption{\textbf{Keyword Spotting (KWS).} Process of recognizing predefined words from a speech signal. The process involves a feature extraction step before the signal is fed into a DNN model. The model then compares the input to the predefined words and detects their presence \cite{Zhang2018}.}
    \label{fig:kws_data}
\end{figure}

\vspace{0.2cm}
\textbf{To illustrate this step of the AI pipeline further, we show an example of data ingestion for a KWS application}. Automatic Speech Recognition (ASR) is a classical AI challenge where data, algorithms, and hardware must be brought together to perform well. Speech source with regional accents has high phonetic variance even in the same word, and expressions in dialects may exceed the standard transcription dictionary. These scenarios suppose a great challenge as retraining a recognizer system with a large amount of real-user data becomes necessary before deploying the system. Therefore, only stakeholders that can acquire custom data and train on it can overcome such a challenge. Keyword Spotting, a particular case of ASR, is the process of recognizing predefined words from a speech signal, which in many cases, serves as a "wake-up" signal to initiate a larger service, see Fig. \ref{fig:kws_data}.

We have set the \textit{AI Data Ingestion} step to download the complete Google Speech Commands dataset \cite{SpeechCommands} containing WAV files. The raw data is downloaded from the provider, parsed, and standardized into HDF5 format files to comply with reusability and compatibility. Because it is a labeled dataset, we skip labeling in the pipeline. Finally, data are partitioned into training, validation, and benchmarking sets. Regarding the training set, the import tool stores each WAV file together with sample ID and class label in one data tensor. This single data tensor is used as input to the next tool. Validation dataset and test dataset can be imported in the same way.

However, raw audio samples are difficult to process. Human's voice is generated from vocal cords vibration, which varies with different sounds. Spectral features of speech signals are more representative than sound waveform in speech recognition. Human's cochlea is more capable of discerning low-frequency signals than high-frequency signals. To model cochlea characteristics, we apply Mel scale over the power spectrum, and it shows higher resolution on low-frequency bands. After discrete cosine transform of Mel log powers, we get Mel-frequency spectral coefficients (MFCCs). MFCC is a widely used audio feature format for speech recognition \cite{1021072}. 

Given the flexibility and modularity of the AI pipeline, we have integrated the \textit{MFCC generation} extraction process into the \textit{Data ingestion step} as pre-processing option. This process employs an audio feature generation tool which produces MFCC features of each audio sample and saves them together with the keyword labels in an HDF5 file. The audio feature generation has been accomplished by leveraging the Librosa library \cite{McFee2015}. Since audio files in Google Speech Command dataset are recorded in 16kHz sampling rate, a moving frame of 128ms length and 32ms stride generates 32 temporal windows in one second. We apply 40 frequency bands per frame, and the output MFCC features of one-second long audio sample are in a 40x32 tensor. The generated MFCC features (training set and test set) can also be reused for training and benchmarking tools of new models. 

\section{TRAINING (2/4)} \label{training_sec}
Training and accuracy benchmarking follows the data ingestion step of the AI pipeline. The training step usually consists of a tool that requires a training dataset (and usually a validation dataset) and produces a model. Since data are standardized, the same training tool can be used with different datasets from the same problem type. On the other hand, the benchmarking tool requires a test dataset and a trained model to produce an accuracy report as output. Visualization tools are often used at this point to help the training process.

In general, an AI pipeline may require a different training framework, another versions of the same one or even several configurations to solve different AI challenges. These differences often lead to software configuration problems when a user tries to solve several AI challenges in the same machine. In this respect, the main benefit of using the Bonseyes AI pipeline for training is the encapsulation of all the needed software and dependencies inside the Docker container that runs the tool. Thus, it is possible to have tools with different deep learning frameworks or different configurations without interfering with each other since they run in an isolated environment. Nonetheless, training procedures are defined in separate training tools to facilitate tools reusability and modularity. Bonseyes includes off-the-shelf Caffe and PyTorch frameworks, but any other framework could also be included.

The flexibility of the dockerized training pipeline allows us to create additional tools that perform model optimizations during training, such as quantization or sparsification. In this case, the new model can be trained from scratch using these optimizations or using a pre-trained model with a new training dataset to optimize and adapt the final model. 


To describe this step of the AI pipeline further, we detail an example of training for a KWS application on Caffe. Two different KWS neural network architectures have been created to cross-compare accuracy, memory usage, and inference time: Convolutional Neural Network (CNN) and Depth-wise Separable CNN (DS-CNN). Since the Long Short-Term Memory (LSTM) -based models do not show a significant advantage of accuracy over DS-CNN, memory footprint and inference latency \cite{Zhang2018}, we only develop two types of CNN models in this paper. In the following subsections, we introduce the training configuration that we have followed, the networks architectures that we have developed and, a final optimization step: a Neural Architecture Search.\par



\subsection{Training Configurations}
All training tools generate both the training model and the solver definition files automatically. We have trained the CNN and DS\_CNN models using Bonseyes-Caffe, \cite{bonseyesgit}. These tools import the output generated in the MFCC generation step using the training dataset where the extracted MFCC features and labels are packed all together into an HDF5 file. Training is carried out with a multinomial logistic loss and Adam optimizer \cite{ADAM} over a batch of 100 MFCC samples (since our input sample size is $40\times32$, we opt to use a relatively big batch size). The batch size and number of iterations are specified in the workflow files that control the execution of the tools. Each model is trained for 40K iterations following a multi-step training strategy. The initial learning rate is $5\times10^{-3}$. With every step of 10K iterations, learning rate drops to 30\% of the previous step.


Further, a benchmarking tool has been built to validate the trained models. This tool takes two inputs: the MFCC features generated from the test dataset (HDF5 file) and the trained model. The inference is performed, and the predicted classes are compared with the provided ground truth, and the results are stored in a JSON file. 

\subsection{Network Architectures}
\label{sec:archi}
To build a KWS model with a small footprint, we have started off by modeling from a 6-layer convolutional neural network. 
The CNN architecture contains 6 convolution layers, 1 average pooling layer, and 1 output layer. More specifically, each convolution layer is followed by 1 batch normalization layer \cite{BatchNorm}, 1 scale layer and 1 ReLU layer \cite{ReLU}. Output features of the pooling layer are flattened to one dimension before connecting with the output layer. CNN model architecture is explained per layer in Table \ref{table:kws_models_0}. We can see that the first convolution layer uses a non-square kernel of size 4 x 10. This kernel maps on 4 rows and 10 columns of an MFCC input image, which in turn refers to 4 frequency bands and 10 sampling windows. A 4 x 10 kernel has an advantage of capturing power variation in a longer period and narrower frequency bands. This setting complies with \cite{Zhang2018}. In the following convolution layers (Conv2 $\sim$ Conv6) 3 x 3 square kernels are applied. 

Depthwise Separable CNN (DS\_CNN) was introduced by \cite{MobileNet}, and we also apply it for KWS. 
DS-CNN improves the execution of standard CNN as it reduces the number of multiplication operations by dividing a standard convolution into two parts: depth-wise and point-wise convolution. In this study, we substitute the standard convolutional layer from the CNN model by depthwise separable convolutions. Thus,
a DS\_CNN model has 1 basic convolution layer (Conv), 5 depthwise separable convolution layers (DS\_Conv), 1 average pooling layer, and 1 output layer. Both parts of the DS\_Conv layer, depthwise convolution, and pointwise convolution, are followed by 1 normalization layer, 1 scale layer, and 1 ReLU layer. The first convolution layer is Conv instead of DS\_Conv. This setting follows the original MobileNet~\cite{MobileNet}, and it helps to extract 2D structure from the input.

\begin{table*}[h]
\centering
\begin{tabular}{rccccccccc}
Model & $conv1^{*}$ & $conv2^{*}$ & conv3 & conv4 & conv5 & conv6 & TOP-1 & $MFP_{ops}$  & Size (KB)  \\
\toprule
CNN & 4x10, 100 & 3x3, 100 & 3x3, 100 & 3x3, 100 & 3x3, 100 & 3x3, 100 & 94.2\% & 581.1 & 1832 \\
\midrule
DS\_CNN & 4x10, 100 & 3x3, 100 & 3x3, 100 & 3x3, 100 & 3x3, 100 & 3x3, 100 & 90.6\% & 69.9 & 1017 \\
\bottomrule
\end{tabular}
\vspace{0.3cm}
\caption{\textbf{Initial architectures of CNN and DS\_CNN networks.} Filter shape is defined as {$k_h\times k_w, M$} where $k_h$ and $k_w$ are kernel height and kernel width, and $M$ is the number of output channels. {*} \textit{conv1} has stride shape $1\times2$ and \textit{conv2} has stride shape $2\times2$. All other convolutional layers have stride shape $1\times1$.}
\label{table:kws_models_0}
\end{table*}


\subsection{Neural Architecture Search}
\label{sec:nas}

Although the manually designed CNN network achieves 94.2\% prediction accuracy (Table \ref{table:kws_models_0}), it is not guaranteed to be the best choice for deployment. Applications on embedded devices are sensitive to energy consumption, and a KWS model is required to respond with low latency. The exact inference latency needs to be gauged on the hardware test, but we can hold an assumption that a model with a smaller number of floating point operations ($FP_{ops}$) will be faster and be more energy efficient. Joint optimization of model accuracy and $FP_{ops}$ is challenging because these two metrics cannot be implemented in one loss function. A solution of model selection comes from Neural Architecture Search (NAS). NAS is a process of automating network architecture engineering~\cite{NAS_survey_2018}. Alternatively speaking, NAS offers a method to automatically explore high-dimensional network hyperparameter space and populate network candidates with good prediction accuracy. \par

Elsken et al. categorized the tasks of NAS in three folds, including search space, search strategy, and performance estimation strategy~\cite{NAS_survey_2018}. In this work, we have applied the popular search strategy Tree-structured Parzen Estimator (TPE)~\cite{bergstra2011algorithms}. The performance estimation strategy that we have followed is a multinomial logistic loss function, and we have used Microsoft NNI library~\cite{MSNNI} for the NAS experiment. The main challenge comes from search space setup as the number of possibilities grows exponentially. We first explore optimization parameters (including learning rate, batch size, weight decay strategy and training iterations) and then we freeze a set of optimization settings to explore the target network parameters: kernel height $k_h$, kernel width $k_w$ and output channel number $M$ of each convolution layer. When the network parameter search space is explored, a selection of CNN architectures is available. These architectures are populated in a two-dimensional space of model accuracy and $FP_{ops}$. We use Pareto frontier~\cite{pareto_1991} to select candidate architectures. If one candidate architecture is on Pareto frontier, it means no other candidate can be more accurate without paying a higher computational cost, and vice versa. We've created an integrated solution of neural architecture search and Pareto frontier selection, for the aim of performance-oriented model selection. More details can be found in~\cite{nas_hpcs2019}.

\section{DEPLOYMENT OPTIMIZATION (3/4)} \label{sec_deploy}
\begin{figure}
    \includegraphics[width=0.9\textwidth]{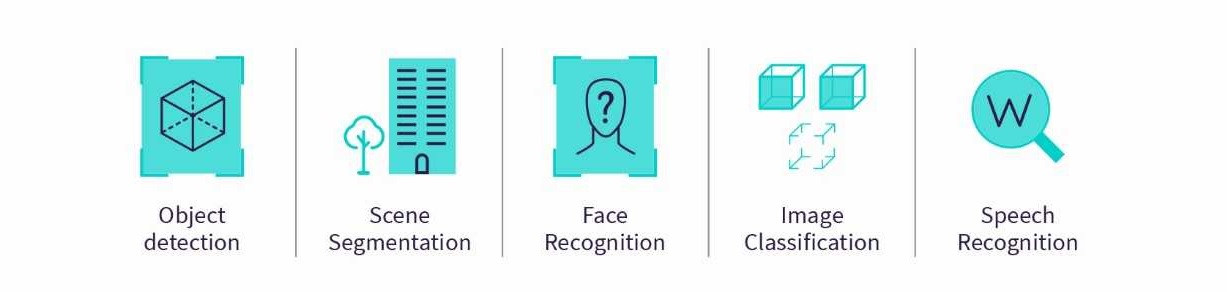}
    \caption{\textbf{AI application types.}}
    \label{fig:ai-app-types}
\end{figure}

\begin{figure}[t]
    \includegraphics[width=0.55\textwidth]{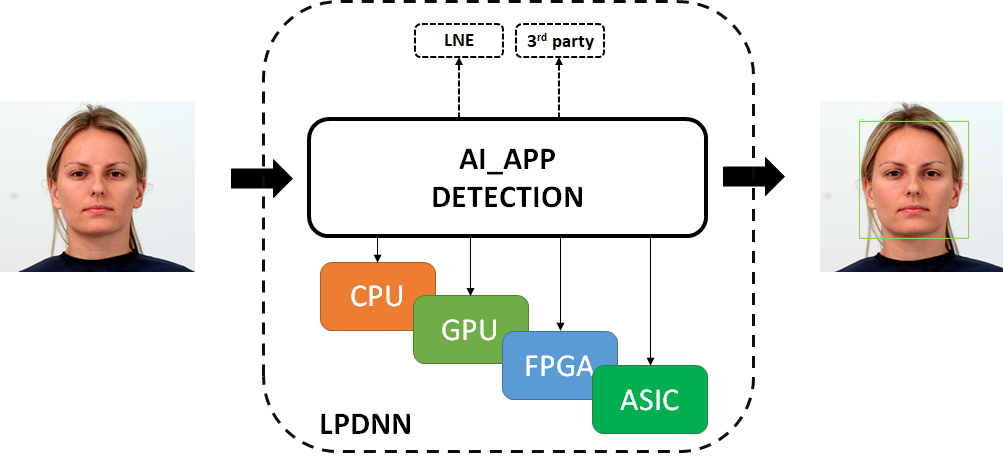}
    \caption{\textbf{Low-power Deep Neural Network (LPDNN) framework.} LPDNN enables the deployment and optimization of AI applications on heterogeneous embedded platforms such as CPU, GPU, GPU and, ASIC.}
    \label{fig:ai-app}
\end{figure}
After training a Deep Neural Network (DNN), the next step in the Bonseyes AI pipeline is the deployment of such DNN on embedded devices. The support and optimization for the deployment of DNNs rely on LPDNN. LPDNN, which stands for Low Power Deep Neural Network, is an enabling deployment framework which provides the tools and capabilities to generate portable and efficient implementations of DNNs for constrained and autonomous applications such as Healthcare Auxiliary, Consumer Emotional Agent, and Automotive Safety and Assistant. The main goal of LPDNN is to provide a set of AI applications, e.g., object detection, image classification, speech recognition, see Fig. \ref{fig:ai-app-types}, which can be deployed and optimized across heterogeneous platforms, e.g., CPU, GPU, FPGA, DSP, ASIC, see Fig. \ref{fig:ai-app}. In this work, we integrate LPDNN into the AI pipeline and present its lightweight architecture and deployment capabilities for embedded devices. Further, we show the deployment of KWS, image classification, and object detection applications on a set of embedded platforms while comparing to other deployment frameworks.

\subsection{LPDNN Architecture}
\begin{figure}
    \includegraphics[width=0.7\textwidth]{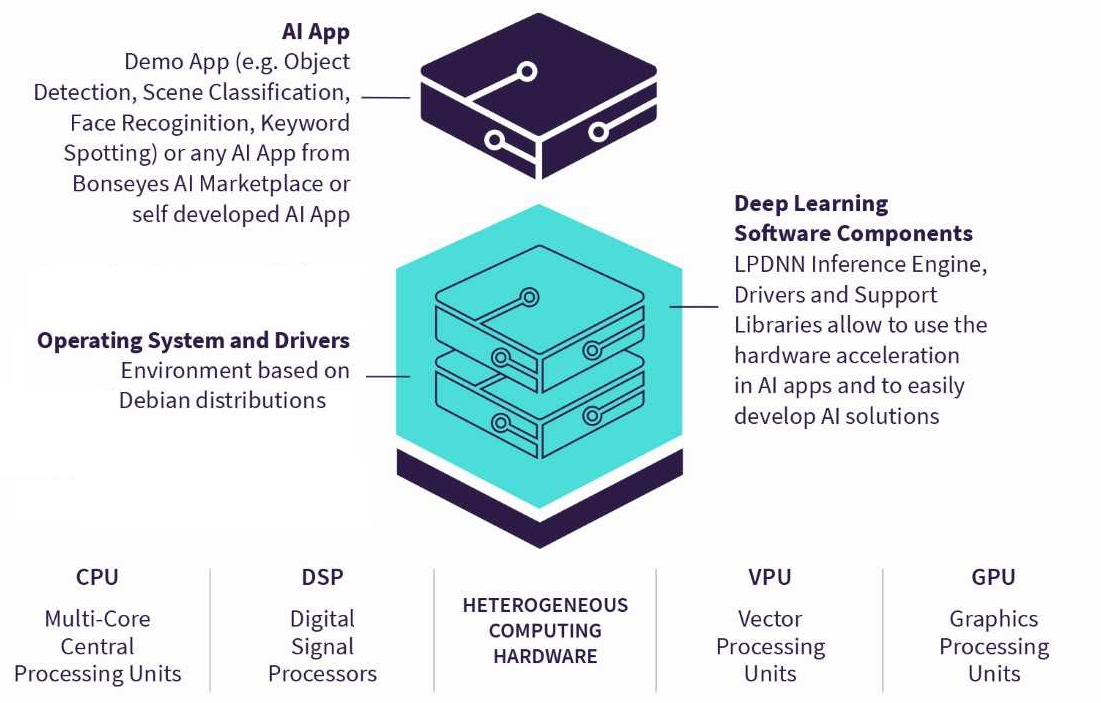}
    \caption
    {\textbf{LPDNN full stack.} LPDNN provides a complete development flow for AI solutions for embedded devices by providing platform support, sample models, optimization tools and integration of external libraries.}
    \label{fig:lpdnn-stack}
\end{figure}

One of the main issues of AI systems is the hardship to replicate results across different systems \cite{replicate}. To solve that issue, LPDNN features a full development flow for AI solutions on embedded devices by providing platform support, sample models, optimization tools, integration of external libraries and benchmarking at several levels of abstraction, see Fig. \ref{fig:lpdnn-stack}. LPDNN's full development flow makes the AI solution very reliable and easy to replicate across systems. Next, we explain LPDNN architecture by describing the concept of AI applications and the LPDNN Inference Engine.

\subsubsection{AI applications}
AI applications are the result of LPDNN's optimization process and the higher level of abstraction for the deployment of DNN on a target platform. They contain all the necessary elements or modules for the execution of DNN. The minimum number of modules that an AI application may contain are two: pre-processing and inference engine modules. More modules can be included to extend the capabilities of the AI application, e.g., connection of several neural networks in a chain fashion. 

Furthermore, AI applications contain a hierarchical but flexible architecture that allows new modules to be integrated within the LPDNN framework through an extendable and straightforward API. For instance, LPDNN supports the integration of 3rd-party self-contained inference engines for AI applications. The AI application could select as a backend LPDNN Inference Engine (LNE) or any other external inference engine integrated into LPDNN, e.g. Renesas e-AI \cite{renesas}, TI-DL \cite{tidl}. The inclusion of external engines also benefits LPDNN as certain embedded platforms provide their own specific and optimized framework to deploy DNNs on them.

\subsubsection{LPDNN Inference Engine} 
\begin{figure}
    \includegraphics[width=0.75\textwidth]{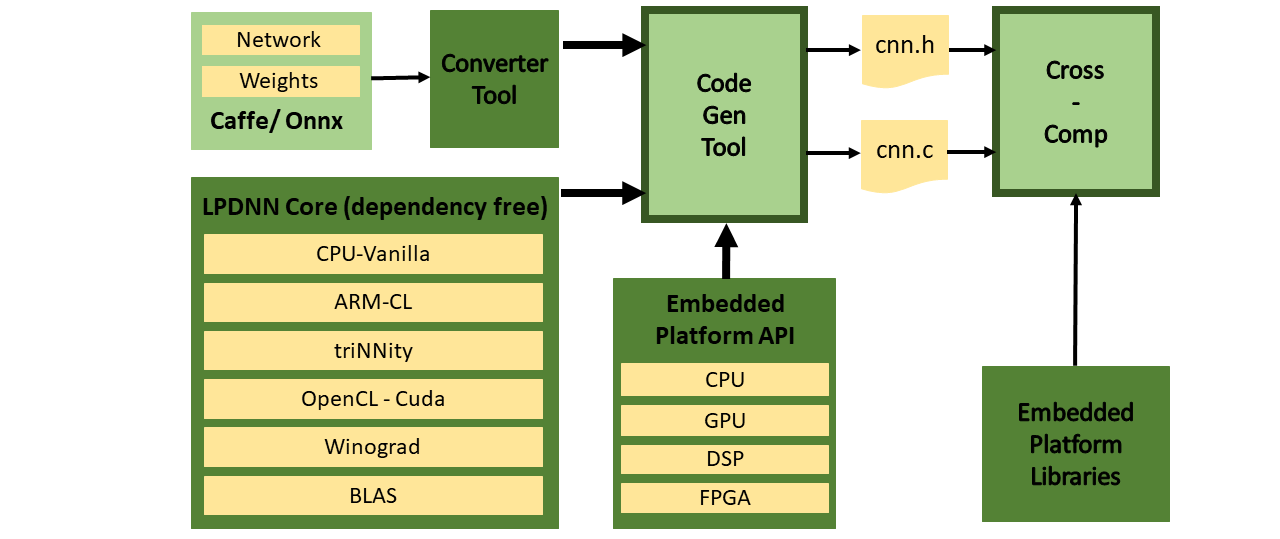}
    \caption{\textbf{LPDNN Inference Engine (LNE) \cite{LPDNN}}.  Plugins can be included for specific layers which allows a broad design space exploration suited for the target platform and  performance specifications.}
    \label{fig:lpdnn}
\end{figure}

In the heart of LPDNN lies the Inference Engine (LNE), initially introduced in \cite{LPDNN}, which is a code generator developed within the Bonseyes project to accelerate the deployment of neural networks on resource-constrained environments\cite{LPDNN}. LNE can generate code for the range of DNN models and across a span of heterogeneous platforms.
LNE supports a wide range of neural network models as it provides direct compatibility with Caffe \cite{Caffe}. In addition, LNE supports ONNX format \cite{onnx}, which allows models trained on any framework to be incorporated into LPDNN providing they can export to ONNX, e.g., PyTorch or TensorFlow. The network model (Caffe, ONNX) is converted to an internal computation graph in an unified format. At this point, several steps such as graph analysis for network compression and memory allocation optimization are performed. LNE provides a plugin-based architecture where a dependency-free inference core is complemented and built together with a set of plugins (acceleration libraries) to produce optimized code for AI applications given a target platform, see Fig. \ref{fig:lpdnn}. Thus, each layer is assigned an implementation among the available computing libraries. Finally, layout conversions are performed in the code generation process to assure the compatibility of the network execution. More details of LNE are provided in \ref{infer_opt}.

\subsubsection{Heterogeneous Computing Support}
One of the main factors for LPDNN's adoption is performance portability across the wide span of hardware platforms. The plugin-based architecture maintains a small and portable core while supporting a wide range of heterogeneous platforms, including CPU, GPU, DSP, and FPGA. One of the objectives of Bonseyes is to provide full support for reference platforms by providing:
\begin{itemize}
  \item Board Support Package (BSP) containing OS images, drivers, and toolchains for several heterogeneous platforms.
  \item A dockerized \cite{docker} and stable environment which increases the reliability by encouraging the replication of results across platforms and environments.
  \item Optimization tools and computing libraries for a variety of computing embedded platforms that can be used by LNE to accelerate the execution of neural networks.
\end{itemize}

\subsection{Inference Optimizations} \label{infer_opt}
LPDNN contains several optimization tools and methods to generate efficient and light implementations for resource-constrained devices.
\subsubsection{Network compression}
LNE supports folding of batch normalization and scale layers into the previous convolution or fully connected layer \cite{bnorm} at compilation time. This optimization provides a reduction in memory size, as the weights of the folded layers are merged, and an acceleration during the inference as the execution of the folded layers are skipped. In addition, some plugins in LNE support fusion of activation layers into the previous convolution at run-time. Fusing activation layers halves the number of memory accesses for a data tensor passing through the combination of convolution + activation layer.
\subsubsection{Memory optimization}
LNE analyzes the computation graph for memory usage and optimizes the overall allocation by sharing the same memory between layers that are not active concurrently (similar to temporary-variables allocation techniques used in compilers). Besides, LNE enables when possible in-place computation: layers share the same memory for input and output tensors.
\subsubsection{Optimized plugins}
\begin{figure}[!htb]
   \begin{minipage}{0.48\textwidth}
     \centering
     \includegraphics[width=\textwidth, height=4cm]{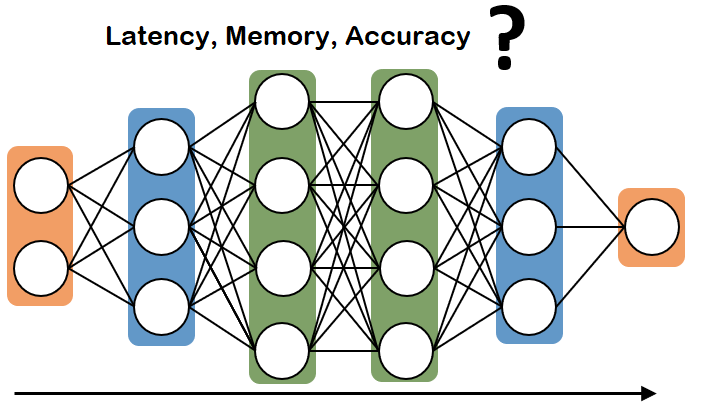}
     \caption{\textbf{Design Space.} Optimization can be achieved by deploying each layer on different processing systems based on their capabilities regarding latency, memory or accuracy. Colours match Fig. \ref{fig:ai-app}.}\label{fig:DSE}
   \end{minipage}\hfill
   \begin{minipage}{0.48\textwidth}
     \centering
    \includegraphics[width=\textwidth, height=3.6cm]{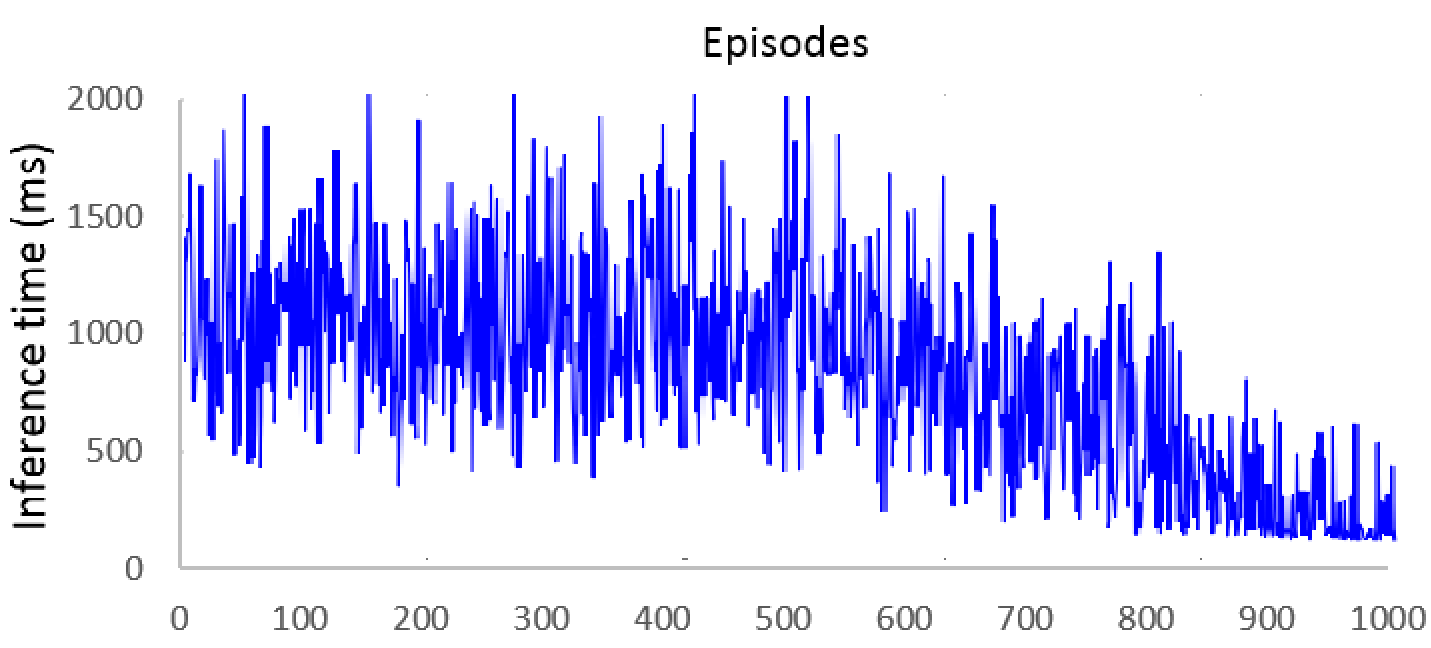}
     \caption{\textbf{RL optimization \cite{RL}}. In a first stage (500 episodes), the agent searches through the design space to learn about the environment. In a second stage, the agent start slowly selecting those implementation that yield a faster inference.}\label{fig:RL}
   \end{minipage}
\end{figure}

Acceleration libraries can be included as plugins in LNE for specific layers to accelerate the inference by calling optimized primitives of the library e.g., BLAS, ARM-CL \cite{armcl}, NNPACK \cite{nnpack} cuDNN. Besides, several libraries can be combined or tuned to boost the performance of the neural network execution. Certain computing processors may provide better performance for specific tasks depending on the model architecture, data type, and layout. Thanks to LNE's flexibility, it is possible to select in what type of processor to deploy each layer and what backend to use, see Fig. \ref{fig:DSE}. Therefore, we find a design exploration problem - named Network Deployment Exploration - when we aim to optimize metrics such as accuracy, latency, or memory.

\subsubsection{Network Deployment Exploration} \label{qs-dnn-section}
To solve this space exploration problem, we propose an automatic exploration framework called QS-DNN, which has been previously introduced in \cite{RL}. QS-DNN implements a learning-based approach, based on Reinforcement Learning \cite{li2017deep}, where an agent explores through the design space, e.g., network deployment across heterogeneous computing processors, and empirically finds an optimal implementation by selecting optimal layer configuration as well as cross-layer optimizations. The network deployment space can be defined as a set of states $\mathcal{S}$, i.e., layer representations. The agent explores such space by employing a set of actions $\mathcal{A}$, i.e., layer implementations, with the final aim of learning a combination of primitives (from LNE's acceleration libraries), that speeds up the performance of the DNN on a given platform, see Fig. \ref{fig:RL}. QS-DNN has been integrated into LPDNN and is tightly coupled with LNE.

\subsubsection{Network Quantization} \label{net_quant}
Neural networks can be further compressed and optimized through approximation \cite{kim2015compression}. A quantization exploration tool has been integrated within LPDNN, which analyzes the sensitivity of each layer to reduced-numerical precision, e.g., int8  \cite{quant}. The tool yields a set of quantization parameters (scale values) which are applied to the weight and output tensors of each layer to minimize the loss in accuracy when using quantized methods. Thanks to LPDNN's benchmark architecture, it is possible to obtain latency measure per layer as well as the accuracy of the network for specific quantization parameters.

\subsection{Deployment of AI applications} \label{deploy_ai}
To demonstrate LPDNN's optimizations and capabilities, we compare LPDNN with several deployment frameworks for KWS, image classification, and object detection applications on a set of embedded platforms. The reader is referred to Section \ref{comparison-embedded-result} for the analytical results of the following comparison scenarios.
\subsubsection{\textbf{LPDNN vs Caffe (KWS)}}
KWS applications are often used as a "wake-up" signal for larger systems due to their low energy consumption, which makes them affordable for always-on modes. Based on this motivation, we focus on deploying the KWS models on a single core of a multi-core CPU platform while leaving available other cores for more consuming tasks or powered-off if not needed. We have chosen the Nvidia Jetson Nano \cite{nano}, which features a quad-core ARM Cortex A-57. Since the KWS models have been trained on Caffe, see Section \ref{training_sec}, we compare the deployment of LPDNN against Caffe for such models. 
\subsubsection{\textbf{LPDNN vs PyTorch (object detection)}} \label{body-pose_sec}
Object detection applications must detect, identify, and localize multiple subjects in the input image. In this work, we show a body-pose estimation model, a case of object detection, where the model has to identify and estimate the skeleton of the people. We take two pre-trained resnet-based models, which have been initially trained by \cite{kreiss2019pifpaf} on PyTorch. We export the model to ONNX and import it in LPDNN where we can leverage LPDNN's optimization and, thus, compare PyTorch deployment with LPDNN's. As body-pose estimation is a very computationally intensive task, we evaluate a heterogeneous implementation for which we propose a last generation automotive platform, the Nvidia Jetson Xavier, which features 8-core ARM v8.2 and a 512-core Volta GPU \cite{xavier}.
\subsubsection{\textbf{Comparison with Embedded Deployment Frameworks (image classification)}} \label{comparison-embedded}
We further compare LPDNN with state-of-the-art deployment frameworks which provide inference engines for resource-constrained devices. Image classification is a classical AI applications where an image needs to be classified according to its content. We show the deployment of a representative range of models for the ImageNet challenge \cite{Imagenet} on two embedded platforms: The Raspberry 3b+ \cite{rpi} and Raspberry 4b \cite{rpi4} featuring a quad-core ARM Cortex-A53 and Cortex-A72, respectively. We especially evaluate several network topologies for resource-constrained devices, e.g., Mobilenets, Squeezenet,  but also reasonably large networks like Resnet50, which allows us to show how the inference engines adapt to the requirements of each network on the selected target platform. We evaluate the following deployment frameworks: \textit{i) Caffe-SSD} \cite{Caffe-ssd}, \textit{ii) ArmCL-DEV20191107} \cite{armcl}, \textit{iii) MNN-0.2.1.5} \cite{mnn}, \textit{iv) NCNN-20191113} \cite{ncnn}, \textit{v) Tengine-DEV20190906} \cite{tengine}, \textit{vi) TFLite-2.0.0} \cite{TFlite}, \textit{vii) LPDNN-20191101.}

\section{IoT HUB INTEGRATION (4/4)}
IoT hub integration is the last step of the AI pipeline. The rise and spread of highly distributed embedded systems bring about scalability issues in the deployment and integration of such systems. Generally, AI applications run on systems which are part of a broader application and service ecosystem that support a value chain. The inclusion of AI enabled systems into an IoT ecosystem is of particular interest when it comes to resource-constrained systems to alleviate computing and memory demands. The heterogeneous nature of IoT and embedded devices not only in terms of HW capabilities, i.e., computation, memory and power consumption, but also in terms of SW support supposes a great challenge to build a global ecosystem. Further, several other problems, such as security and privacy, need to be addressed in a distributed platform where information is continuously shared.  

\subsection{IoT scenarios}
\begin{figure}
    \includegraphics[width=0.6\textwidth]{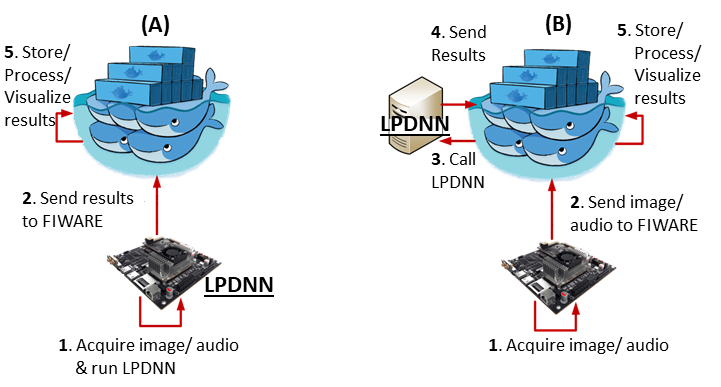}
    \caption{\textbf{Bonseyes IoT tools.} Bonseyes relies on the FIWARE platform \cite{fiware} for the implementation of the integration pipeline. It backs two main scenarios A) Edge-processing, B) Cloud-processing.}
    \label{fig:fiware}
\end{figure}

The IoT hub integration in the AI pipeline backs two main scenarios which are relevant for low-power and autonomous environments, e.g., command voice in medical health care, pedestrian detection in automotive, etc:

\begin{itemize}
    \item \textbf{Edge-processing}: Data are processed on the embedded device, and results are retrieved and stored in the cloud for further processing and exploitation, see Fig. \ref{fig:fiware}-A. 
    \item \textbf{Cloud-processing} Part of the computation steps in the embedded system is delegated to server- or cloud-based AI engines. This scenario can be of great interest for constrained systems when their resources are not able to offer enough computational power to execute AI algorithms, see Fig. \ref{fig:fiware}-B. 
\end{itemize}


\subsection{Bonseyes IoT Tools}
Bonseyes relies on the FIWARE platform \cite{fiware} for the implementation of the IoT hub integration. Both \textit{edge-processing} and \textit{cloud-processing} scenarios are supported by the use of a set of FIWARE Generic Enablers not only to exchange data between different enablers but also to manage embedded systems as IoT agents. Also, the second scenario requires the use of Kurento Media Server to seamlessly transfer media contents from the embedded platforms to the cloud computing infrastructure.

\subsubsection{FIWARE}
It is an open-source community which provides a rich set of APIs to facilitate the connection to IoT devices, user interaction, and process of data. FIWARE offers a rich library of components, called Generic Enablers (GE), which provide reference implementations that allow users to develop new applications. GEs provide the general-purpose functions such as data context management, IoT service enablement, advance web-based UI, security, interface to networks, the architecture of application/services ecosystem, and cloud hosting.

\subsubsection{Kurento Media Server}
Kurento is a Stream-Oriented GE providing a media server and a set of APIs to help the development of web-based media applications for browsers or smartphones. It offers ready-made bricks of media processing algorithms such as computer vision, augmented reality, and speech analysis. \par

\vspace{0.2cm}In this work, we have focused on the \textit{edge processing} scenario, since the selected embedded platforms in Section \ref{sec_deploy} are fairly able to process data directly on the \textit{Edge}. Hence, we have created a dedicated media module in Kurento which call LPDNN's AI application and stores the results.

\section{{Results and discussion}}
In this section, we introduce the results of the AI pipeline and detail the outcome of the \textit{Training} and \textit{Deployment} steps. We give examples of different AI applications while we analyze and demonstrate the effectiveness and benefits of the Bonseyes AI pipeline for embedded systems.
\subsection{Training}
Table \ref{table:benchmarks} shows benchmark scores of CNN models and DS\_CNN models trained with Bonseyes-Caffe \cite{bonseyesgit} for the KWS application shown in Section \ref{training_sec}. The test set contains 2567 audio samples which are recorded from totally different speakers of the training samples. After 40K iterations training, the CNN model marks 94.23\% accuracy on the test set, and the model size is 1.8 MB. DS\_CNN model marks 90.65\% accuracy, and the model size is 1 MB. With current hyper-parameter settings, DS\_CNN is 4\% less accurate than CNN, but its model size is about half of CNN. According to Zhang et al.\cite{Zhang2018}, DS\_CNN has the potential to be more accurate, and we will refine this model. \par

In this work, we have decided to build small footprint KWS models to ease the deployment on embedded devices. As introduced in Section \ref{training_sec}, we can apply quantization (\textit{Q}), and sparsification (\textit{S}) functions to obtain a further compression in the models. On both CNN and DS\_CNN, \textit{Q} and \textit{S} have minor disadvantage ($<0.7\%$ loss) on test accuracy. Moreover, 16-bit fixed-point quantization can save half memory space and reduced bandwidth requirements at run time. An \textit{S} model may also leverage the amount of zeroes in its matrices and obtain benefits in memory and computation when it is deployed. Finally, We also observe that a \textit{Q}+\textit{S} model is more accurate than an \textit{S} model as quantization may act as a regularizer and slightly increase the accuracy.

\begin{table}[tp]
\begin{center}
  \begin{tabular}{ | l | c | c | c || l | c | c | c |}
    \hline
    Model & Acc & Sparsity & Size (KB) & Model & Acc & Sparsity & Size (KB)\\ \hline
    CNN & 94.23\% & 0\% & 1832 & DS\_CNN & 90.65\%  & 0\% & 1017 \\ 
    CNN + \textit{Q} & 94.04\% & 0\% & 918 & DS\_CNN + \textit{Q} & 90.62\%  & 0\%  & 511     \\ 
    CNN + \textit{S} & 93.69\% & 39.6\% &  1832 & DS\_CNN + \textit{S} & 89.96\% & 27.9\% & 1017  \\ 
    CNN + \textit{Q} + \textit{S} & 94.27\% & 39.8\% &  918 & DS\_CNN + \textit{Q} + \textit{S} & 90.19\% & 27.7\% & 511   \\ \hline
  \end{tabular}
\end{center}
\vspace{0.2cm}
\caption{\textbf{KWS trained models:} Benchmark of the models on test set. \textit{Q}: Quantization (16-bit), \textit{S}: Sparsity.}
\label{table:benchmarks}
\vspace{-0.5cm}
\end{table}

Manually designed CNN and DS\_CNN models achieved over 90\% prediction accuracy on KWS (Table \ref{table:kws_models_0}). However, these models contain much redundancy and suppose a challenge for performance on embedded devices. We propose Neural Architecture Search method to explore KWS models with reduced model size and $FP_{ops}$. 12 CNN models are spotted through NAS and Pareto-optimal selection~\cite{nas_hpcs2019} and 3 models are presented in Table \ref{tab:pareto_best_CNN}. In \textit{kws1}, we can see that kernel sizes of \textit{conv2} to \textit{conv6} are no longer fixed at $3\times3$ but vary from $1\times1$ to $5\times5$. Output channels are all below 50. All these modifications reduce $MFP_{ops}$ from 581.1 to 223.4 and improve TOP-1 accuracy from 94.2\% to 95.1\%. Further model size reduction is found in \textit{kws3} and \textit{kws9} by the price of a minor drop in accuracy. Another observation is that $4\times10$ kernels of the first convolution layer are no longer needed for an accurate KWS model. These rectangular kernels were designed to cover a longer temporal sequence than frequency bands from MFCC features~\cite{Zhang2018}. As MFCC features were generated with 128ms frame length in this study, much longer than 40ms in literature, CNN with only square kernels is capable of delivering accurate KWS models.

Overall, we can see that NAS discovers obsolete network sub-structures introduced by manual network design. We leverage the advantages of DS\_CNN and adapt CNN architectures in Table \ref{tab:pareto_best_CNN} to DS\_CNN version (Table \ref{tab:pareto_best_DSCNN}). Three new DS\_CNN architectures are coined and are trained in 300K iterations. Each DS\_CNN model achieves higher prediction accuracy than the seed DS\_CNN model. In the meantime, the $MFP_{ops}$ are only 11.9, 9.7 and 7.0 respectively. \textit{ds\_kws9} reports the minimum computational load in this study.

\subsection{Deployment Optimization}
Following the description of Section \ref{deploy_ai}, we employ LPDNN's tools to optimize the deployment of AI applications on the embedded platforms, and we compare it with several well-known deployment frameworks. To ensure a fair comparison across the frameworks, we have enabled all the optimizations as provided by each vendor of the framework. All benchmarks have been performed identically: calculating the average of ten inferences after an initial (discarded) warm-up run. In addition, we have set the platforms in performance mode, which sets the clocks to the highest frequency. To ensure that the platform does not overheat, triggering thermal throttling, we have monitored the platforms and sampled the OS registers each second.

\subsubsection{\textbf{LPDNN vs Caffe (KWS)}}
\begin{figure}
\centering
\begin{subfigure}{0.45\linewidth}
  \centering
     \includegraphics[width=\linewidth]{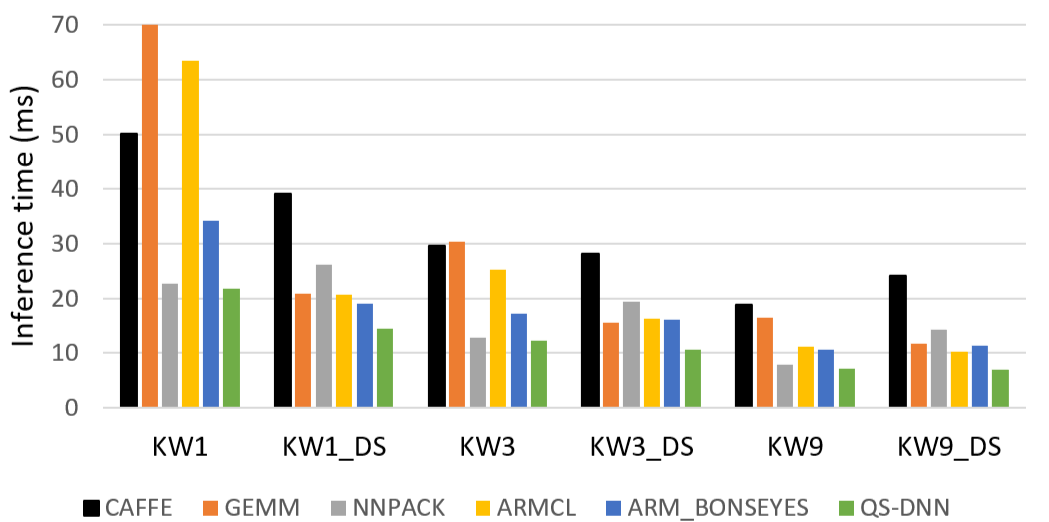}
     \caption{Inference time using FP32 operations. Caffe's time is given in black. Other colours represent the deployment of LPDNN employing a single library. On green, QS-DNN's solution after having learnt an optimized combination of primitives.}\label{fig:kws_results}
\end{subfigure}
\hspace{0.2cm}
\begin{subfigure}{0.45\linewidth}
  \centering
  \includegraphics[width=\linewidth]{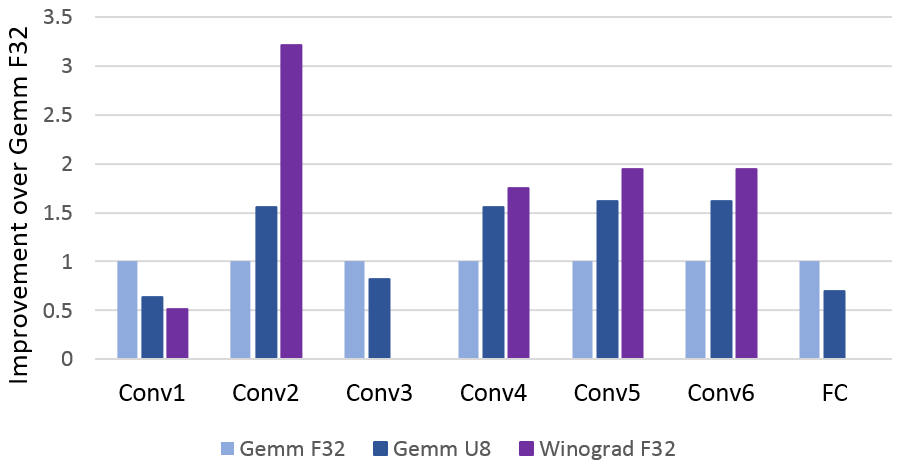}
  \caption{Quantization analysis for KWS1. Speedup of GEMM int8 primitives over GEMM F32 and comparison with Winograd F32 for each layer of the network.}
  \label{fig:kws-quant}
\end{subfigure}
\caption{\textbf{LPDNN vs Caffe (KWS).} Inference time using single-thread operations on the Nvidia Jetson Nano (CPU) (the lower the better).}
\label{fig:kws-app}
\end{figure}

We compare the deployment of the trained KWS models from Table \ref{tab:pareto_best_CNN} and \ref{tab:pareto_best_DSCNN} between Caffe and LPDNN on the Nvidia Jetson Nano platform. Caffe has been installed natively on the platform using Openblas as backend. On the LPDNN side, we employ LNE coupled with an RL-based search (QS-DNN) to find an optimized solution for the deployment on the target platform. Fig. \ref{fig:kws_results} shows the results of the deployment when processing a one-second audio input using a single-thread and 32-bit floating-point operations on the CPU.

Overall, we can see that while Caffe takes, from largest to smallest, between 50~ms and 24~ms to process a single KWS network while LPDNN employs between 21~ms and 7~ms. Caffe-Openblas featuring general-matrix multiplication (GEMM) only outperforms LPDNN-GEMM on KWS1. Nonetheless, QS-DNN's capability to learn an optimized combination of libraries makes LPDNN significantly outperform Caffe on every network being up to x3.5 faster. Regarding LPDNN's acceleration libraries, it is noted that no single library outperforms all other libraries across all networks. QS-DNN, however, always outperforms all individual libraries across all networks. Hence, it is possible to prove that QS-DNN adapts to each specific use case and supposes a powerful tool for LPDNN's optimized deployment.

Networks can be further compressed through quantization by employing primitives featuring reduced-numerical precision, as explained in Section \ref{net_quant}. Fig. \ref{fig:kws-quant} illustrates an analysis of KWS1's layers using int8 primitives from armCL. We observe that GEMM int8 generally - but not always - outperforms its FP32 counterpart. The overall improvement of having KWS1 full int8 accounts for 52\% over GEMM FP32, as Conv4-Conv6 are the most computational intensive layers, and $1/4$ of the memory size with only 1\% drop in accuracy. However, this improvement is shadowed by efficient convolution primitives such as Winograd \cite{winograd}, whose F32 implementation outperforms GEMM FP32 by 88\%. Based on these results, we state that the use of quantization on Jetson Nano devices provides a trade-off between latency and memory consumption for the KWS application.

\subsubsection{\textbf{LPDNN vs PyTorch (object detection)}}
\begin{figure}
\centering
\begin{subfigure}{.5\textwidth}
  \centering
  \includegraphics[width=0.7\linewidth]{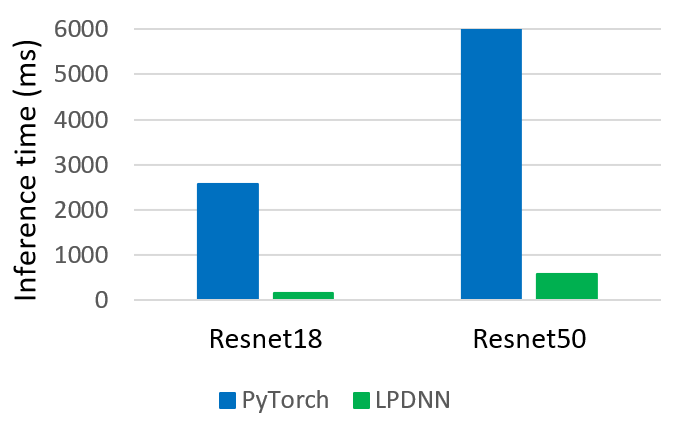}
  \caption{CPU deployment using single-thread FP32.}
  \label{fig:body-pose-sub1}
\end{subfigure}%
\begin{subfigure}{.5\textwidth}
  \centering
  \includegraphics[width=0.7\linewidth]{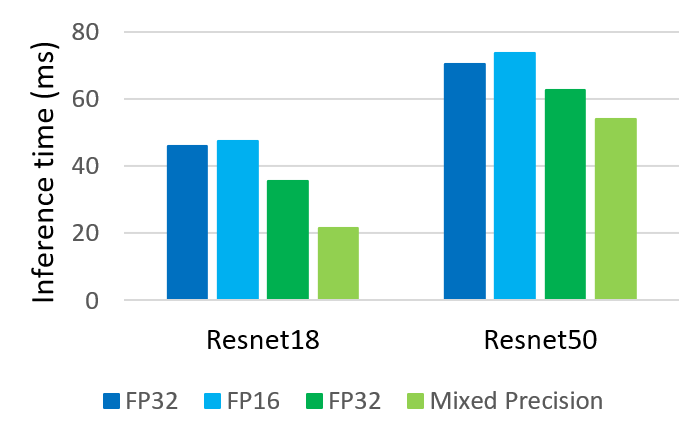}
  \caption{GPU deployment using FP32 and FP16.}
  \label{fig:body-pose-sub2}
\end{subfigure}
\caption{\textbf{LPDNN vs PyTorch (Object detection).} Inference time of the (resnet-based) body-pose estimation models on the Nvidia Jetson Xavier.}
\label{fig:body-pose_results}
\end{figure}

We compare the deployment of the (resnet-based) body-pose estimation models presented in Section \ref{body-pose_sec} between PyTorch and LPDNN on the Nvidia Xavier platform. We have installed PyTorch natively on the platform using the latest release provided by Nvidia \cite{pytorch-nvidia}. We perform a first experiment deploying the models on the Arm CPU of the platform and employing single-thread and 32-bit floating-point operations. PyTorch uses the ATen library based on c++11, while LPDNN employs QS-DNN coupled to LNE to find an optimized combination of primitives. Fig. \ref{fig:body-pose-sub1} presents the deployment results, and we can observe that LPDNN amply outperforms PyTorch on the CPU, being up to x15 faster for the resnet18-based model.

As PyTorch is mainly a training framework, we assume that CPU has not been largely optimized in favour of GPU deployment. Hence, we perform a second experiment on the GPU employing the same backend, CUDA-10, in both frameworks. As it can be in Fig. \ref{fig:body-pose-sub2}, LPDNN outperforms PyTorch on both networks performing up to 28\% faster. Further, we analyze the performance of half-precision to speed up the inference and reduce memory footprint. PyTorch employing FP16 out-of-the-box turns out to be slower than FP32. This might be due to a direct conversion from FP32 to FP16, and, as \cite{half} suggests, this conversion needs to be carefully carried out to keep performance up. In LPDNN, by contrast, we set QS-DNN to automatically learn what data type performs better and give an optimized combination of primitives for LNE. Thus, we achieve up to 65\% improvement on Resnet18 when leveraging mixed precision.

\subsubsection{\textbf{Comparison with Embedded Deployment Frameworks (image classification)}} \label{comparison-embedded-result}

To demonstrate the capabilities of LPDNN further, we compare it with a range of embedded deployment frameworks on the RPI3b+ and RPI4b\footnote{Both platforms have been flashed with 64-bit Debian OS images} for the ImageNet challenge as stated in Section \ref{comparison-embedded}. We have built all the deployment frameworks natively and enabled all the optimizations as provided by each vendor. To further guarantee a fair comparison, we have chosen five representative networks that are used across all deployment frameworks: \textit{Alexnet}, \textit{Resnet50-V1}, \textit{Googlenet-V1}, \textit{Squeezenet-V1.1}, and \textit{Mobilenet-V2-1.0-224}. The reference networks are taken from Caffe repository, which we assume as the reference framework, and are imported into each of the tested frameworks directly if supported by the framework, or via an official conversion tool (provided by the framework\footnote{With exception of TF Lite. We convert Caffe to TF via MMDNN and ONNX. From TF to TF Lite we use the official converter.}). All inferences are performed using a single-thread and 32-bit floating-point operations on the CPU.

\begin{figure}
\centering
\begin{subfigure}{.5\textwidth}
  \centering
  \includegraphics[width=\linewidth]{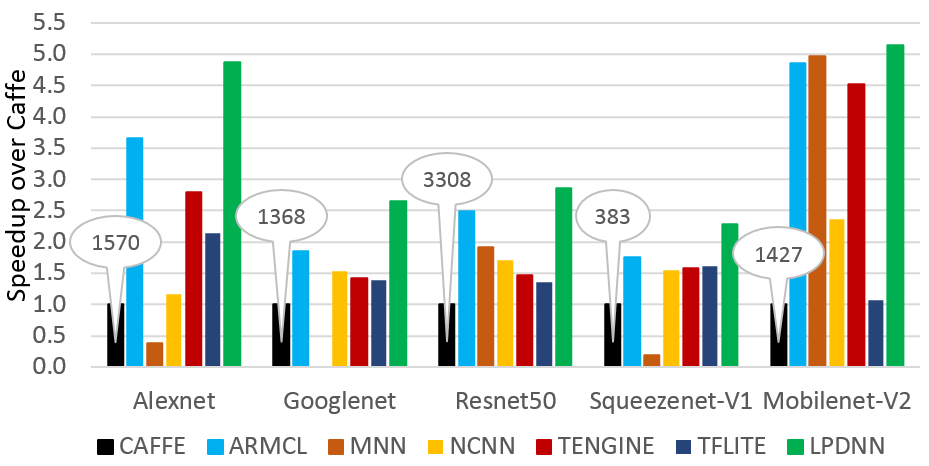}
  \caption{RPI3}
  \label{fig:engines-rpi3}
\end{subfigure}%
\begin{subfigure}{.5\textwidth}
  \centering
  \includegraphics[width=\linewidth]{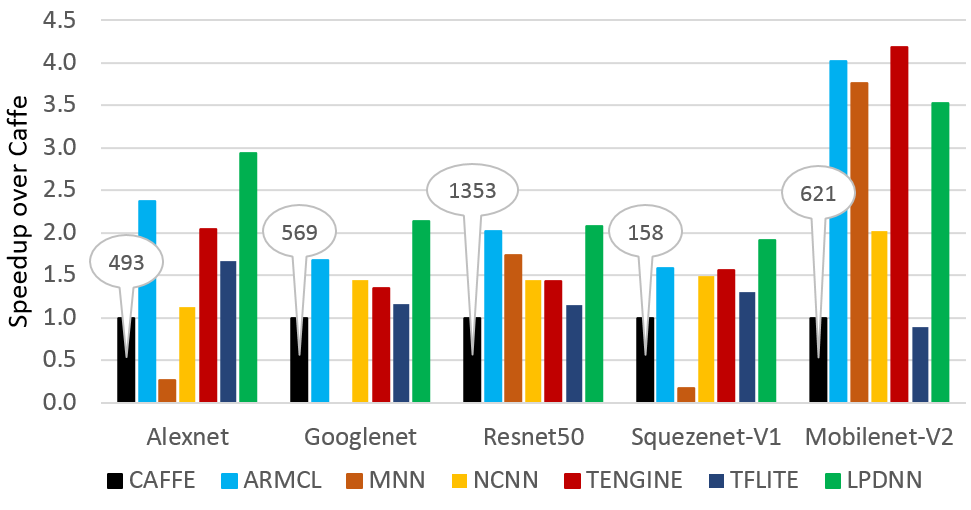}
  \caption{RPI4}
  \label{fig:engines-rpi4}
\end{subfigure}
\caption{\textbf{Comparison with embedded deployment frameworks (Image classification).} Inference results of the range of deployment frameworks for the reference networks on the RPI3 and RPI4. The bars represent the relative speedup over Caffe, which displays the absolute time in milliseconds.}
\label{fig:engine_comparison_results}
\end{figure}

Fig.~\ref{fig:engine_comparison_results} presents the results of the reference networks across the range of deployment frameworks on the RPI3 and RPI4. We show the relative speedup of each framework with respect to Caffe (reference) for which we display the absolute time in milliseconds. 
We can easily observe two general trends. \textit{i)} Certain frameworks perform very well on a single network but drastically drop performance on other networks, e.g., MNN, Tengine. \textit{ii)} the deployment of some networks has been remarkably optimized, e.g., several frameworks achieve over 4x improvement over the reference in Mobilenet-V2, while generally, no framework accomplishes a comparable speed-up in other network topologies, e.g., Googlenet, Squeezenet.

These two trends confirm the importance of selecting several network topologies to have a sound estimation of the  frameworks and how they can adapt to each structure. 
ArmCL and LPDNN are the frameworks that provide the most stable performance improvements. From these two, LPDNN obtains the highest speedups and outperforms all other frameworks across networks and target platforms performing over 2x better than the average and 5x better than the worst performing framework. LPDNN's high performance can be explained by the abundant number of optimized primitives that LPDNN contains and its ability to learn a combination of primitives for each network. The stability across the two platforms proves that LPDNN is robust and can adapt to different architectures while retaining high performance.

\textbf{TF Lite exception.}
We have benchmarked all frameworks taking the networks from Caffe as reference. TF Lite is the only framework that neither supports nor provides an official conversion tool for Caffe networks and hence, we have employed MMDNN and ONNX for the conversion to TF. From TF to TF Lite we use the official converter. We argue that this conversion, although providing correct output classifications results, might be the cause of the low performance of TF Lite. Therefore, we offer a new benchmark comparing TF Lite with LPDNN taking three networks from TF repositories instead. We take Mobilenet-V2 from TF Lite repository and Googlenet-V1 and Resnet50-V1 from the original TF as they are not available in TF Lite directly. We convert all of them from TF to LPDNN via ONNX and the last two from TF to TF Lite via TF Lite official converter. Table \ref{tab:tf_table} depicts the results of such benchmarks on the RPI3 and RPI4.

\begin{table}[]
    \centering
    \begin{tabular}{|c|c|c|c|c|}
        \hline 
        DNN & \multicolumn{2}{c|}{RPI3} & \multicolumn{2}{c|}{RPI4}\\
        \cline{2-5}  &  LPDNN & TF Lite & LPDNN & TF Lite  \\
        \hline 
        Mobilenet-V2 (from TF Lite) & 217 & 246 & 105 & 119 \\ 
        Googlenet (from TF) & 429 & 839 & 216 & 430 \\ 
        Resnet50 (from TF) & 1172 & 2024 & 667 & 981 \\ 
        \hline 
    \end{tabular}
    \vspace{0.2cm}
    \caption{Inference comparison in milliseconds between TF Lite and LPDNN taking TF original networks.}
    \label{tab:tf_table}
    \vspace{-0.8cm}
\end{table}

We can remark that the native TF Lite network, Mobilenet-V2, performs notably well, achieving almost the performance of LPDNN. However, we can already see that the original networks from standard TF, converted to TF Lite, drop in performance being up to 2.5x slower than LPDNN. This point denotes the lack of proper support in TensorFlow for other formats, e.g., Caffe, ONNX, TF, as it appears that TF Lite only performs well when the networks have been written in a specific format, i.e., TF Lite format, or contains a specific architecture, e.g., Mobilenet. This fact supposes a constrained for users wanting to employ custom models as they would either have poor performance when executing non TF Lite networks or have TF framework a fixed dependency. Nevertheless, LPDNN also outperforms TF Lite using TF original models. We can thus prove LPDNN's flexibility and support for other network's formats and its domain over the range of embedded-oriented deployment frameworks.

\section{Conclusion and Future Work} 
Nowadays, training and deployment of custom AI solutions on embedded and IoT devices poses many issues as it requires a fine-grained integration of data, algorithms, and tools. These barriers prevent the massive spread of AI applications in our daily life as only end-to-end systems can overcome these hurdles and achieve accurate and fast solutions. In this work, we present a modular end-to-end AI pipeline architecture, which brings data, algorithms, and deployment tools together to facilitate the production and porting of AI solution for embedded devices. We ease the integration and lower the required expertise by providing key benefits such as the reusability of tools thanks to a dockerized API, and the flexibility to add new steps to the workflow. Thus, we propose a pipeline with four main steps: i) data ingestion, ii) model training, iii) deployment optimization and, iv) the IoT hub integration. 

We have demonstrated the effectiveness of the AI pipeline by providing several examples of AI applications in each of the steps and show the significance of a tight integration of the pipeline steps toward having an efficient and competitive implementation. Thus, we are able to create a data ingestion step for the Google speech commands dataset seamlessly and train two families of CNN and DS\_CNN networks achieving up to 95.1\% and 92.6\%. Further, we have presented the lightweight architecture and deployment capabilities of our deployment framework, LPDNN, and demonstrate that it outperforms all other popular deployment frameworks on a set of AI applications and across a range of embedded platforms.

As future work, we envision to fully optimize and deploy trained models on very low-power devices that can be employed for applications such as healthcare sensing, wearable systems, or car-driving assistant.

\begin{acks}
This project has received funding from the European Union's Horizon 2020 research and innovation programme under grant agreement No 732204 (Bonseyes). This work is supported by the Swiss State Secretariat for Education, Research and Innovation (SERI) under contract number 16.0159. The opinions expressed and arguments employed herein do not necessarily reflect the official views of these funding bodies.
\end{acks}

\bibliographystyle{ACM-Reference-Format}
\bibliography{sample-base}


\begin{thebibliography}{76}


\ifx \showCODEN    \undefined \def \showCODEN     #1{\unskip}     \fi
\ifx \showDOI      \undefined \def \showDOI       #1{#1}\fi
\ifx \showISBNx    \undefined \def \showISBNx     #1{\unskip}     \fi
\ifx \showISBNxiii \undefined \def \showISBNxiii  #1{\unskip}     \fi
\ifx \showISSN     \undefined \def \showISSN      #1{\unskip}     \fi
\ifx \showLCCN     \undefined \def \showLCCN      #1{\unskip}     \fi
\ifx \shownote     \undefined \def \shownote      #1{#1}          \fi
\ifx \showarticletitle \undefined \def \showarticletitle #1{#1}   \fi
\ifx \showURL      \undefined \def \showURL       {\relax}        \fi
\providecommand\bibfield[2]{#2}
\providecommand\bibinfo[2]{#2}
\providecommand\natexlab[1]{#1}
\providecommand\showeprint[2][]{arXiv:#2}

\bibitem[\protect\citeauthoryear{??}{Spe}{2017}]%
        {SpeechCommands}
 \bibinfo{year}{2017}\natexlab{}.
\newblock \bibinfo{title}{Google Speech Commands dataset}.
\newblock
\newblock
\urldef\tempurl%
\url{https://ai.googleblog.com/2017/08/launching-speech-commands-dataset.html}
\showURL{%
\tempurl}


\bibitem[\protect\citeauthoryear{??}{onn}{2017a}]%
        {onnx_1}
 \bibinfo{year}{2017}\natexlab{a}.
\newblock \bibinfo{title}{The ONNX project}.
\newblock
\newblock
\urldef\tempurl%
\url{https://github.com/onnx/onnx}
\showURL{%
\tempurl}


\bibitem[\protect\citeauthoryear{??}{onn}{2017b}]%
        {onnx}
 \bibinfo{year}{2017}\natexlab{b}.
\newblock \bibinfo{title}{Open Neural Network Exchange (ONNX)}.
\newblock
\newblock
\urldef\tempurl%
\url{https://onnx.ai/}
\showURL{%
\tempurl}


\bibitem[\protect\citeauthoryear{??}{arm}{2018}]%
        {armcl}
 \bibinfo{year}{2018}\natexlab{}.
\newblock \bibinfo{title}{Arm Compute Library}.
\newblock
\newblock
\urldef\tempurl%
\url{https://developer.arm.com/ip-products/processors/machine-learning/compute-library}
\showURL{%
\tempurl}


\bibitem[\protect\citeauthoryear{??}{bon}{2018}]%
        {bonseyesgit}
 \bibinfo{year}{2018}\natexlab{}.
\newblock \bibinfo{title}{Bonseyes Official Caffe 1.0 Version}.
\newblock
\newblock
\urldef\tempurl%
\url{https://github.com/bonseyes/caffe-jacinto}
\showURL{%
\tempurl}


\bibitem[\protect\citeauthoryear{??}{Caf}{2018a}]%
        {Caffe}
 \bibinfo{year}{2018}\natexlab{a}.
\newblock \bibinfo{title}{Caffe}.
\newblock
\newblock
\urldef\tempurl%
\url{http://caffe.berkeleyvision.org/}
\showURL{%
\tempurl}


\bibitem[\protect\citeauthoryear{??}{Caf}{2018b}]%
        {Caffe-ssd}
 \bibinfo{year}{2018}\natexlab{b}.
\newblock \bibinfo{title}{Caffe-SSD}.
\newblock
\newblock
\urldef\tempurl%
\url{https://github.com/weiliu89/caffe}
\showURL{%
\tempurl}


\bibitem[\protect\citeauthoryear{??}{ICT}{2018}]%
        {ICT}
 \bibinfo{year}{2018}\natexlab{}.
\newblock \bibinfo{title}{Discover the power of Artificial Intelligence to
  drive ICT innovation}.
\newblock
\newblock
\urldef\tempurl%
\url{https://news.itu.int/discover-the-power-of-artificial-intelligence-to-drive-ict-innovation-in-the-first-issue-of-the-itu-journal/}
\showURL{%
\tempurl}


\bibitem[\protect\citeauthoryear{??}{doc}{2018}]%
        {docker}
 \bibinfo{year}{2018}\natexlab{}.
\newblock \bibinfo{title}{Docker}.
\newblock
\newblock
\urldef\tempurl%
\url{http://www.docker.com}
\showURL{%
\tempurl}


\bibitem[\protect\citeauthoryear{??}{fiw}{2018}]%
        {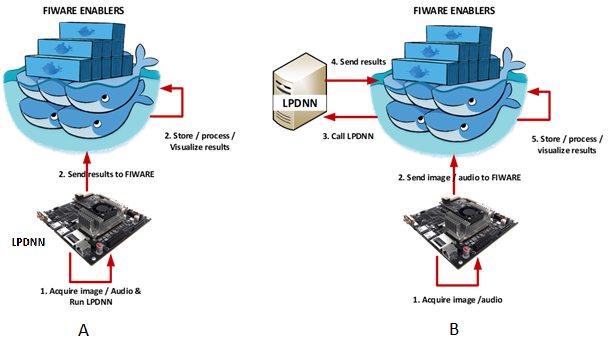}
 \bibinfo{year}{2018}\natexlab{}.
\newblock \bibinfo{title}{FI-ware Project}.
\newblock
\newblock
\urldef\tempurl%
\url{https://www.fiware.org/}
\showURL{%
\tempurl}


\bibitem[\protect\citeauthoryear{??}{fra}{2018}]%
        {fragmentation}
 \bibinfo{year}{2018}\natexlab{}.
\newblock \bibinfo{title}{Machine Learning Fragmentation Is Slowing Us Down:
  There Is a Solution}.
\newblock
\newblock
\urldef\tempurl%
\url{https://www.cmswire.com/digital-experience/machine-learning-fragmentation-is-slowing-us-down-there-is-a-solution/}
\showURL{%
\tempurl}


\bibitem[\protect\citeauthoryear{??}{rep}{2018}]%
        {replicate}
 \bibinfo{year}{2018}\natexlab{}.
\newblock \bibinfo{title}{Scientists Can't Replicate AI Studies. That's Bad
  News}.
\newblock
\newblock
\urldef\tempurl%
\url{https://futurism.com/scientists-cant-replicate-ai-studies}
\showURL{%
\tempurl}


\bibitem[\protect\citeauthoryear{??}{ama}{2019}]%
        {amazon_ml}
 \bibinfo{year}{2019}\natexlab{}.
\newblock \bibinfo{title}{Amazon Machine Learning on AWS}.
\newblock
\newblock
\urldef\tempurl%
\url{https://aws.amazon.com/machine-learning}
\showURL{%
\tempurl}


\bibitem[\protect\citeauthoryear{??}{sag}{2019}]%
        {sagemaker}
 \bibinfo{year}{2019}\natexlab{}.
\newblock \bibinfo{title}{Amazon SageMaker}.
\newblock
\newblock
\urldef\tempurl%
\url{https://aws.amazon.com/sagemaker/}
\showURL{%
\tempurl}


\bibitem[\protect\citeauthoryear{??}{NNA}{2019}]%
        {NNAPI}
 \bibinfo{year}{2019}\natexlab{}.
\newblock \bibinfo{title}{Android Neural Networks API (NNAPI)}.
\newblock
\newblock
\urldef\tempurl%
\url{https://developer.android.com/ndk/guides/neuralnetworks}
\showURL{%
\tempurl}


\bibitem[\protect\citeauthoryear{??}{app}{2019}]%
        {apple}
 \bibinfo{year}{2019}\natexlab{}.
\newblock \bibinfo{title}{Apple AI}.
\newblock
\newblock
\urldef\tempurl%
\url{https://www.zdnet.com/article/apple-says-artificial-intelligence-and-machine-learning-critical-area-as-it-promotes-ai-chief/}
\showURL{%
\tempurl}


\bibitem[\protect\citeauthoryear{??}{caf}{2019a}]%
        {caffe2}
 \bibinfo{year}{2019}\natexlab{a}.
\newblock \bibinfo{title}{CAFFE2}.
\newblock
\newblock
\urldef\tempurl%
\url{https://caffe2.ai/}
\showURL{%
\tempurl}


\bibitem[\protect\citeauthoryear{??}{caf}{2019b}]%
        {caffe2fb}
 \bibinfo{year}{2019}\natexlab{b}.
\newblock \bibinfo{title}{CAFFE2FB}.
\newblock
\newblock
\urldef\tempurl%
\url{https://research.fb.com/downloads/caffe2/}
\showURL{%
\tempurl}


\bibitem[\protect\citeauthoryear{??}{CNT}{2019}]%
        {CNTK}
 \bibinfo{year}{2019}\natexlab{}.
\newblock \bibinfo{title}{CNTK: The Microsoft Cognitive Toolkit}.
\newblock
\newblock
\urldef\tempurl%
\url{https://docs.microsoft.com/en-us/cognitive-toolkit}
\showURL{%
\tempurl}


\bibitem[\protect\citeauthoryear{??}{Cor}{2019a}]%
        {CoreML_conv}
 \bibinfo{year}{2019}\natexlab{a}.
\newblock \bibinfo{title}{Converting Trained Models to Core ML}.
\newblock
\newblock
\urldef\tempurl%
\url{https://developer.apple.com/documentation/coreml/converting_trained_models_to_core_ml}
\showURL{%
\tempurl}


\bibitem[\protect\citeauthoryear{??}{Cor}{2019b}]%
        {CoreML}
 \bibinfo{year}{2019}\natexlab{b}.
\newblock \bibinfo{title}{Core ML: Integrate machine learning models into your
  app}.
\newblock
\newblock
\urldef\tempurl%
\url{https://developer.apple.com/documentation/coreml}
\showURL{%
\tempurl}


\bibitem[\protect\citeauthoryear{??}{bno}{2019}]%
        {bnorm}
 \bibinfo{year}{2019}\natexlab{}.
\newblock \bibinfo{title}{Folding of bnorm into convolution}.
\newblock
\newblock
\urldef\tempurl%
\url{https://tehnokv.com/posts/fusing-batchnorm-and-conv/}
\showURL{%
\tempurl}


\bibitem[\protect\citeauthoryear{??}{goo}{2019a}]%
        {google_ai}
 \bibinfo{year}{2019}\natexlab{a}.
\newblock \bibinfo{title}{Google AI Platform}.
\newblock
\newblock
\urldef\tempurl%
\url{https://cloud.google.com/ai-platform/}
\showURL{%
\tempurl}


\bibitem[\protect\citeauthoryear{??}{goo}{2019b}]%
        {google_ai_tools}
 \bibinfo{year}{2019}\natexlab{b}.
\newblock \bibinfo{title}{Google AI Tools}.
\newblock
\newblock
\urldef\tempurl%
\url{https://ai.google/tools/}
\showURL{%
\tempurl}


\bibitem[\protect\citeauthoryear{??}{goo}{2019c}]%
        {google}
 \bibinfo{year}{2019}\natexlab{c}.
\newblock \bibinfo{title}{Google ML}.
\newblock
\newblock
\urldef\tempurl%
\url{https://cloud.google.com/products/machine-learning/}
\showURL{%
\tempurl}


\bibitem[\protect\citeauthoryear{??}{Gre}{2019a}]%
        {Greengrass}
 \bibinfo{year}{2019}\natexlab{a}.
\newblock \bibinfo{title}{Greengrass}.
\newblock
\newblock
\urldef\tempurl%
\url{https://aws.amazon.com/greengrass/}
\showURL{%
\tempurl}


\bibitem[\protect\citeauthoryear{??}{Gre}{2019b}]%
        {GreengrassRegion}
 \bibinfo{year}{2019}\natexlab{b}.
\newblock \bibinfo{title}{Greengrass Region}.
\newblock
\newblock
\urldef\tempurl%
\url{https://aws.amazon.com/about-aws/global-infrastructure/regional-product-services/}
\showURL{%
\tempurl}


\bibitem[\protect\citeauthoryear{??}{Ima}{2019}]%
        {Imagenet}
 \bibinfo{year}{2019}\natexlab{}.
\newblock \bibinfo{title}{ImageNet}.
\newblock
\newblock
\urldef\tempurl%
\url{URL: http://www.image-net.org}
\showURL{%
\tempurl}


\bibitem[\protect\citeauthoryear{??}{Ope}{2019}]%
        {OpenVINO}
 \bibinfo{year}{2019}\natexlab{}.
\newblock \bibinfo{title}{Intel OpenVINO Toolkit}.
\newblock
\newblock
\urldef\tempurl%
\url{https://docs.openvinotoolkit.org}
\showURL{%
\tempurl}


\bibitem[\protect\citeauthoryear{??}{Ker}{2019}]%
        {Keras}
 \bibinfo{year}{2019}\natexlab{}.
\newblock \bibinfo{title}{Keras: The Python Deep Learning Library}.
\newblock
\newblock
\urldef\tempurl%
\url{https://keras.io}
\showURL{%
\tempurl}


\bibitem[\protect\citeauthoryear{??}{lea}{2019}]%
        {lead}
 \bibinfo{year}{2019}\natexlab{}.
\newblock \bibinfo{title}{Lead in AI}.
\newblock
\newblock
\urldef\tempurl%
\url{https://www.forbes.com/sites/danielaraya/2019/01/01/who-will-lead-in-the-age-of-artificial-intelligence/#4f3a15aa6f95}
\showURL{%
\tempurl}


\bibitem[\protect\citeauthoryear{??}{azu}{2019a}]%
        {azure_ml}
 \bibinfo{year}{2019}\natexlab{a}.
\newblock \bibinfo{title}{Microsoft Azure}.
\newblock
\newblock
\urldef\tempurl%
\url{https://azure.microsoft.com/en-us/services/machine-learning/}
\showURL{%
\tempurl}


\bibitem[\protect\citeauthoryear{??}{azu}{2019b}]%
        {azure_iot}
 \bibinfo{year}{2019}\natexlab{b}.
\newblock \bibinfo{title}{Microsoft Azure IoT Edge}.
\newblock
\newblock
\urldef\tempurl%
\url{https://azure.microsoft.com/en-in/services/iot-edge/}
\showURL{%
\tempurl}


\bibitem[\protect\citeauthoryear{??}{mnn}{2019}]%
        {mnn}
 \bibinfo{year}{2019}\natexlab{}.
\newblock \bibinfo{title}{Mobile Neural Network (MNN): a lightweight deep
  neural network inference engine}.
\newblock
\newblock
\urldef\tempurl%
\url{https://github.com/alibaba/MNN}
\showURL{%
\tempurl}


\bibitem[\protect\citeauthoryear{??}{azu}{2019c}]%
        {azure_interpret}
 \bibinfo{year}{2019}\natexlab{c}.
\newblock \bibinfo{title}{Model interpretability in Azure Machine Learning
  service}.
\newblock
\newblock
\urldef\tempurl%
\url{https://docs.microsoft.com/en-us/azure/machine-learning/service/how-to-machine-learning-interpretability}
\showURL{%
\tempurl}


\bibitem[\protect\citeauthoryear{??}{MXN}{2019}]%
        {MXNet}
 \bibinfo{year}{2019}\natexlab{}.
\newblock \bibinfo{title}{MXNet: A Flexible and Efficient Library for Deep
  Learning}.
\newblock
\newblock
\urldef\tempurl%
\url{https://mxnet.apache.org}
\showURL{%
\tempurl}


\bibitem[\protect\citeauthoryear{??}{ncn}{2019}]%
        {ncnn}
 \bibinfo{year}{2019}\natexlab{}.
\newblock \bibinfo{title}{NCNN: A High-Performance Neural Network Inference
  Framework Optimized for the Mobile Platform.}
\newblock
\newblock
\urldef\tempurl%
\url{https://github.com/Tencent/ncnn}
\showURL{%
\tempurl}


\bibitem[\protect\citeauthoryear{??}{nnp}{2019}]%
        {nnpack}
 \bibinfo{year}{2019}\natexlab{}.
\newblock \bibinfo{title}{NNPACK}.
\newblock
\newblock
\urldef\tempurl%
\url{https://github.com/Maratyszcza/NNPACK}
\showURL{%
\tempurl}


\bibitem[\protect\citeauthoryear{??}{nan}{2019}]%
        {nano}
 \bibinfo{year}{2019}\natexlab{}.
\newblock \bibinfo{title}{Nvidia Jetson Nano}.
\newblock
\newblock
\urldef\tempurl%
\url{https://www.nvidia.com/en-us/autonomous-machines/embedded-systems/jetson-nano/}
\showURL{%
\tempurl}


\bibitem[\protect\citeauthoryear{??}{xav}{2019}]%
        {xavier}
 \bibinfo{year}{2019}\natexlab{}.
\newblock \bibinfo{title}{Nvidia Jetson Xavier}.
\newblock
\newblock
\urldef\tempurl%
\url{https://developer.nvidia.com/embedded/jetson-agx-xavier-developer-kit}
\showURL{%
\tempurl}


\bibitem[\protect\citeauthoryear{??}{pyt}{2019}]%
        {pytorch-nvidia}
 \bibinfo{year}{2019}\natexlab{}.
\newblock \bibinfo{title}{Nvidia realeased of PyTorch}.
\newblock
\newblock
\urldef\tempurl%
\url{https://docs.nvidia.com/deeplearning/frameworks/pytorch-release-notes/overview.html#overview}
\showURL{%
\tempurl}


\bibitem[\protect\citeauthoryear{??}{Pyt}{2019}]%
        {Pytorch}
 \bibinfo{year}{2019}\natexlab{}.
\newblock \bibinfo{title}{PyTorch: From Research to Production}.
\newblock
\newblock
\urldef\tempurl%
\url{https://pytorch.org}
\showURL{%
\tempurl}


\bibitem[\protect\citeauthoryear{??}{qua}{2019}]%
        {quant}
 \bibinfo{year}{2019}\natexlab{}.
\newblock \bibinfo{title}{Quantization analysis tool}.
\newblock
\newblock
\urldef\tempurl%
\url{https://github.com/BUG1989/caffe-int8-convert-tools}
\showURL{%
\tempurl}


\bibitem[\protect\citeauthoryear{??}{rpi}{2019a}]%
        {rpi}
 \bibinfo{year}{2019}\natexlab{a}.
\newblock \bibinfo{title}{Raspberry Pi3}.
\newblock
\newblock
\urldef\tempurl%
\url{https://www.raspberrypi.org/products/raspberry-pi-3-model-b-plus/}
\showURL{%
\tempurl}


\bibitem[\protect\citeauthoryear{??}{rpi}{2019b}]%
        {rpi4}
 \bibinfo{year}{2019}\natexlab{b}.
\newblock \bibinfo{title}{Raspberry Pi4}.
\newblock
\newblock
\urldef\tempurl%
\url{https://www.raspberrypi.org/products/raspberry-pi-4-model-b/}
\showURL{%
\tempurl}


\bibitem[\protect\citeauthoryear{??}{ren}{2019}]%
        {renesas}
 \bibinfo{year}{2019}\natexlab{}.
\newblock \bibinfo{title}{Renesas e-AI}.
\newblock
\newblock
\urldef\tempurl%
\url{https://www.renesas.com/eu/en/solutions/key-technology/e-ai.html}
\showURL{%
\tempurl}


\bibitem[\protect\citeauthoryear{??}{ten}{2019a}]%
        {tengine}
 \bibinfo{year}{2019}\natexlab{a}.
\newblock \bibinfo{title}{Tengine: a Lite, High-performance, and Modular
  Inference Engine for Embedded Device}.
\newblock
\newblock
\urldef\tempurl%
\url{https://github.com/OAID/Tengine}
\showURL{%
\tempurl}


\bibitem[\protect\citeauthoryear{??}{TF}{2019}]%
        {TF}
 \bibinfo{year}{2019}\natexlab{}.
\newblock \bibinfo{title}{TensorFlow: An end-to-end open source machine
  learning platform}.
\newblock
\newblock
\urldef\tempurl%
\url{https://www.tensorflow.org}
\showURL{%
\tempurl}


\bibitem[\protect\citeauthoryear{??}{TFl}{2019}]%
        {TFlite}
 \bibinfo{year}{2019}\natexlab{}.
\newblock \bibinfo{title}{TensorFlow Lite: Deploy machine learning models on
  mobile and IoT devices}.
\newblock
\newblock
\urldef\tempurl%
\url{https://www.tensorflow.org/lite}
\showURL{%
\tempurl}


\bibitem[\protect\citeauthoryear{??}{ten}{2019b}]%
        {tensorrt}
 \bibinfo{year}{2019}\natexlab{b}.
\newblock \bibinfo{title}{TensorRT}.
\newblock
\newblock
\urldef\tempurl%
\url{https://developer.nvidia.com/tensorrt}
\showURL{%
\tempurl}


\bibitem[\protect\citeauthoryear{??}{tid}{2019}]%
        {tidl}
 \bibinfo{year}{2019}\natexlab{}.
\newblock \bibinfo{title}{TI-DL}.
\newblock
\newblock
\urldef\tempurl%
\url{https://training.ti.com/texas-instruments-deep-learning-tidl-overview}
\showURL{%
\tempurl}


\bibitem[\protect\citeauthoryear{??}{Tor}{2019}]%
        {Torch}
 \bibinfo{year}{2019}\natexlab{}.
\newblock \bibinfo{title}{Torch: A Scientific Computing Framework for LuaJIT}.
\newblock
\newblock
\urldef\tempurl%
\url{http://torch.ch}
\showURL{%
\tempurl}


\bibitem[\protect\citeauthoryear{Anderson, Su, Dahyot, and Gregg}{Anderson
  et~al\mbox{.}}{2019}]%
        {nas_hpcs2019}
\bibfield{author}{\bibinfo{person}{Andrew Anderson}, \bibinfo{person}{Jing Su},
  \bibinfo{person}{Rozenn Dahyot}, {and} \bibinfo{person}{David Gregg}.}
  \bibinfo{year}{2019}\natexlab{}.
\newblock \showarticletitle{Performance-Oriented Neural Architecture Search}.
  In \bibinfo{booktitle}{\emph{2019 International Conference on High
  Performance Computing and Simulation}}. IEEE.
\newblock


\bibitem[\protect\citeauthoryear{Bergstra, Bardenet, Bengio, and
  K{\'e}gl}{Bergstra et~al\mbox{.}}{2011}]%
        {bergstra2011algorithms}
\bibfield{author}{\bibinfo{person}{James~S Bergstra}, \bibinfo{person}{R{\'e}mi
  Bardenet}, \bibinfo{person}{Yoshua Bengio}, {and} \bibinfo{person}{Bal{\'a}zs
  K{\'e}gl}.} \bibinfo{year}{2011}\natexlab{}.
\newblock \showarticletitle{Algorithms for hyper-parameter optimization}. In
  \bibinfo{booktitle}{\emph{Advances in neural information processing
  systems}}. \bibinfo{pages}{2546--2554}.
\newblock


\bibitem[\protect\citeauthoryear{contributors{{)}}}{contributors{{)}}}{2019}]%
        {MSNNI}
\bibfield{author}{\bibinfo{person}{{{(}}{{Microsoft NNI}} contributors{{)}}}.}
  \bibinfo{year}{{{2019}}}\natexlab{}.
\newblock \bibinfo{title}{{{An open source AutoML toolkit for neural
  architecture search and hyper-parameter tuning.}}}
\newblock
  \bibinfo{howpublished}{\url{{(}}{{{{https://github.com/Microsoft/nni}}}}{{)}}}.
\newblock
\newblock
\shownote{[Online; accessed 27 May 2019].}


\bibitem[\protect\citeauthoryear{de~Prado, Denna, Benini, and Pazos}{de~Prado
  et~al\mbox{.}}{2018a}]%
        {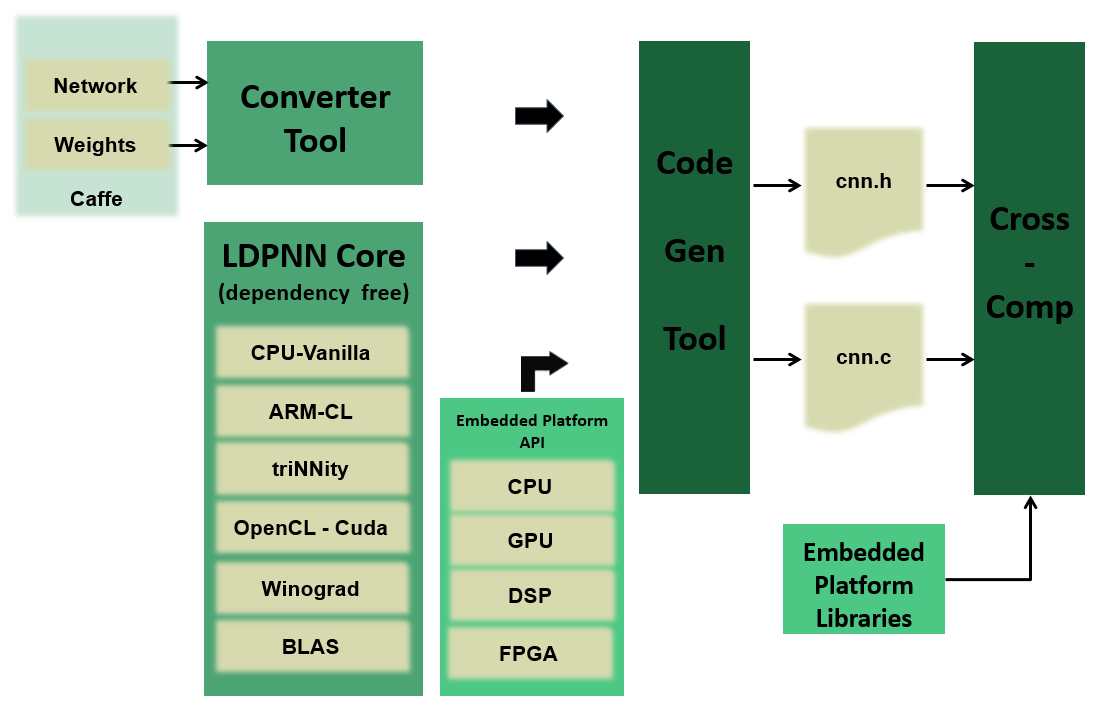}
\bibfield{author}{\bibinfo{person}{Miguel de Prado}, \bibinfo{person}{Maurizio
  Denna}, \bibinfo{person}{Luca Benini}, {and} \bibinfo{person}{Nuria Pazos}.}
  \bibinfo{year}{2018}\natexlab{a}.
\newblock \showarticletitle{QUENN: QUantization engine for low-power neural
  networks}. In \bibinfo{booktitle}{\emph{Proceedings of the 15th ACM
  International Conference on Computing Frontiers}}. ACM,
  \bibinfo{pages}{36--44}.
\newblock


\bibitem[\protect\citeauthoryear{de~Prado, Pazos, and Benini}{de~Prado
  et~al\mbox{.}}{2018b}]%
        {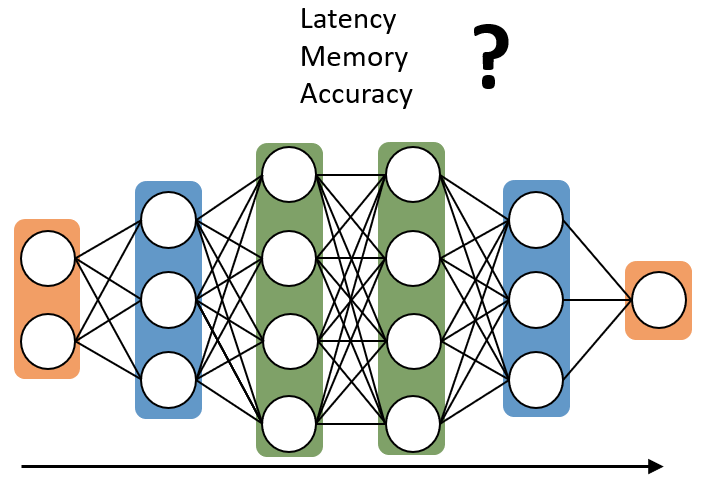}
\bibfield{author}{\bibinfo{person}{Miguel de Prado}, \bibinfo{person}{Nuria
  Pazos}, {and} \bibinfo{person}{Luca Benini}.}
  \bibinfo{year}{2018}\natexlab{b}.
\newblock \showarticletitle{Learning to infer: RL-based search for DNN
  primitive selection on Heterogeneous Embedded Systems}.
\newblock \bibinfo{journal}{\emph{arXiv preprint arXiv:1811.07315}}
  (\bibinfo{year}{2018}).
\newblock


\bibitem[\protect\citeauthoryear{{Elsken}, {Hendrik Metzen}, and
  {Hutter}}{{Elsken} et~al\mbox{.}}{2018}]%
        {NAS_survey_2018}
\bibfield{author}{\bibinfo{person}{Thomas {Elsken}}, \bibinfo{person}{Jan
  {Hendrik Metzen}}, {and} \bibinfo{person}{Frank {Hutter}}.}
  \bibinfo{year}{2018}\natexlab{}.
\newblock \showarticletitle{{Neural Architecture Search: A Survey}}.
\newblock \bibinfo{journal}{\emph{arXiv e-prints}}, Article
  \bibinfo{articleno}{arXiv:1808.05377} (\bibinfo{date}{Aug}
  \bibinfo{year}{2018}), \bibinfo{numpages}{arXiv:1808.05377}~pages.
\newblock
\showeprint[arxiv]{stat.ML/1808.05377}


\bibitem[\protect\citeauthoryear{et~al.}{et~al.}{2017}]%
        {Bonseyes}
\bibfield{author}{\bibinfo{person}{T.~Llewellynn et al.}}
  \bibinfo{year}{2017}\natexlab{}.
\newblock \showarticletitle{BONSEYES: platform for open development of systems
  of artificial intelligence}. In \bibinfo{booktitle}{\emph{Proceedings of the
  Computing Frontiers Conference}}. ACM, \bibinfo{pages}{299--304}.
\newblock


\bibitem[\protect\citeauthoryear{Graves, Mohamed, and Hinton}{Graves
  et~al\mbox{.}}{2013}]%
        {graves2013speech}
\bibfield{author}{\bibinfo{person}{Alex Graves}, \bibinfo{person}{Abdel-rahman
  Mohamed}, {and} \bibinfo{person}{Geoffrey Hinton}.}
  \bibinfo{year}{2013}\natexlab{}.
\newblock \showarticletitle{Speech recognition with deep recurrent neural
  networks}. In \bibinfo{booktitle}{\emph{2013 IEEE international conference on
  acoustics, speech and signal processing}}. IEEE, \bibinfo{pages}{6645--6649}.
\newblock


\bibitem[\protect\citeauthoryear{Ho and Wong}{Ho and Wong}{2017}]%
        {half}
\bibfield{author}{\bibinfo{person}{Nhut-Minh Ho} {and}
  \bibinfo{person}{Weng-Fai Wong}.} \bibinfo{year}{2017}\natexlab{}.
\newblock \showarticletitle{Exploiting half precision arithmetic in Nvidia
  GPUs}. In \bibinfo{booktitle}{\emph{2017 IEEE High Performance Extreme
  Computing Conference (HPEC)}}. IEEE, \bibinfo{pages}{1--7}.
\newblock


\bibitem[\protect\citeauthoryear{Howard, Zhu, Chen, Kalenichenko, Wang, Weyand,
  Andreetto, and Adam}{Howard et~al\mbox{.}}{2017}]%
        {MobileNet}
\bibfield{author}{\bibinfo{person}{Andrew~G. Howard}, \bibinfo{person}{Menglong
  Zhu}, \bibinfo{person}{Bo Chen}, \bibinfo{person}{Dmitry Kalenichenko},
  \bibinfo{person}{Weijun Wang}, \bibinfo{person}{Tobias Weyand},
  \bibinfo{person}{Marco Andreetto}, {and} \bibinfo{person}{Hartwig Adam}.}
  \bibinfo{year}{2017}\natexlab{}.
\newblock \showarticletitle{MobileNets: Efficient Convolutional Neural Networks
  for Mobile Vision Applications}.
\newblock \bibinfo{journal}{\emph{CoRR}}  \bibinfo{volume}{abs/1704.04861}
  (\bibinfo{year}{2017}).
\newblock


\bibitem[\protect\citeauthoryear{Huval, Wang, Tandon, Kiske, Song,
  Pazhayampallil, Andriluka, Rajpurkar, Migimatsu, Cheng-Yue,
  et~al\mbox{.}}{Huval et~al\mbox{.}}{2015}]%
        {huval2015empirical}
\bibfield{author}{\bibinfo{person}{Brody Huval}, \bibinfo{person}{Tao Wang},
  \bibinfo{person}{Sameep Tandon}, \bibinfo{person}{Jeff Kiske},
  \bibinfo{person}{Will Song}, \bibinfo{person}{Joel Pazhayampallil},
  \bibinfo{person}{Mykhaylo Andriluka}, \bibinfo{person}{Pranav Rajpurkar},
  \bibinfo{person}{Toki Migimatsu}, \bibinfo{person}{Royce Cheng-Yue},
  {et~al\mbox{.}}} \bibinfo{year}{2015}\natexlab{}.
\newblock \showarticletitle{An empirical evaluation of deep learning on highway
  driving}.
\newblock \bibinfo{journal}{\emph{arXiv preprint arXiv:1504.01716}}
  (\bibinfo{year}{2015}).
\newblock


\bibitem[\protect\citeauthoryear{Ioffe and Szegedy}{Ioffe and Szegedy}{2015}]%
        {BatchNorm}
\bibfield{author}{\bibinfo{person}{Sergey Ioffe} {and}
  \bibinfo{person}{Christian Szegedy}.} \bibinfo{year}{2015}\natexlab{}.
\newblock \showarticletitle{Batch Normalization: Accelerating Deep Network
  Training by Reducing Internal Covariate Shift}. In
  \bibinfo{booktitle}{\emph{Proceedings of the 32nd International Conference on
  Machine Learning - Volume 37}} \emph{(\bibinfo{series}{ICML'15})}.
  \bibinfo{publisher}{JMLR.org}, \bibinfo{pages}{448--456}.
\newblock
\urldef\tempurl%
\url{http://dl.acm.org/citation.cfm?id=3045118.3045167}
\showURL{%
\tempurl}


\bibitem[\protect\citeauthoryear{Kim, Park, Yoo, Choi, Yang, and Shin}{Kim
  et~al\mbox{.}}{2015}]%
        {kim2015compression}
\bibfield{author}{\bibinfo{person}{Yong-Deok Kim}, \bibinfo{person}{Eunhyeok
  Park}, \bibinfo{person}{Sungjoo Yoo}, \bibinfo{person}{Taelim Choi},
  \bibinfo{person}{Lu Yang}, {and} \bibinfo{person}{Dongjun Shin}.}
  \bibinfo{year}{2015}\natexlab{}.
\newblock \showarticletitle{Compression of deep convolutional neural networks
  for fast and low power mobile applications}.
\newblock \bibinfo{journal}{\emph{arXiv preprint arXiv:1511.06530}}
  (\bibinfo{year}{2015}).
\newblock


\bibitem[\protect\citeauthoryear{Kingma and Ba}{Kingma and Ba}{2015}]%
        {ADAM}
\bibfield{author}{\bibinfo{person}{Diederik~P. Kingma} {and}
  \bibinfo{person}{Jimmy Ba}.} \bibinfo{year}{2015}\natexlab{}.
\newblock \showarticletitle{Adam: {A} Method for Stochastic Optimization}. In
  \bibinfo{booktitle}{\emph{Proceedings of the 3rd International Conference on
  Learning Representations (ICLR 2015)}}.
\newblock


\bibitem[\protect\citeauthoryear{Kreiss, Bertoni, and Alahi}{Kreiss
  et~al\mbox{.}}{2019}]%
        {kreiss2019pifpaf}
\bibfield{author}{\bibinfo{person}{Sven Kreiss}, \bibinfo{person}{Lorenzo
  Bertoni}, {and} \bibinfo{person}{Alexandre Alahi}.}
  \bibinfo{year}{2019}\natexlab{}.
\newblock \showarticletitle{PifPaf: Composite Fields for Human Pose
  Estimation}. In \bibinfo{booktitle}{\emph{The IEEE Conference on Computer
  Vision and Pattern Recognition (CVPR)}}.
\newblock


\bibitem[\protect\citeauthoryear{Li}{Li}{2017}]%
        {li2017deep}
\bibfield{author}{\bibinfo{person}{Yuxi Li}.} \bibinfo{year}{2017}\natexlab{}.
\newblock \showarticletitle{Deep reinforcement learning: An overview}.
\newblock \bibinfo{journal}{\emph{arXiv preprint arXiv:1701.07274}}
  (\bibinfo{year}{2017}).
\newblock


\bibitem[\protect\citeauthoryear{Maji, Mundy, Dasika, Beu, Mattina, and
  Mullins}{Maji et~al\mbox{.}}{2019}]%
        {winograd}
\bibfield{author}{\bibinfo{person}{Partha Maji}, \bibinfo{person}{Andrew
  Mundy}, \bibinfo{person}{Ganesh Dasika}, \bibinfo{person}{Jesse Beu},
  \bibinfo{person}{Matthew Mattina}, {and} \bibinfo{person}{Robert Mullins}.}
  \bibinfo{year}{2019}\natexlab{}.
\newblock \showarticletitle{Efficient Winograd or Cook-Toom Convolution Kernel
  Implementation on Widely Used Mobile CPUs}.
\newblock \bibinfo{journal}{\emph{arXiv preprint arXiv:1903.01521}}
  (\bibinfo{year}{2019}).
\newblock


\bibitem[\protect\citeauthoryear{Mathur}{Mathur}{1991}]%
        {pareto_1991}
\bibfield{author}{\bibinfo{person}{Vijay~K. Mathur}.}
  \bibinfo{year}{1991}\natexlab{}.
\newblock \showarticletitle{How Well Do We Know Pareto Optimality?}
\newblock \bibinfo{journal}{\emph{The Journal of Economic Education}}
  \bibinfo{volume}{22}, \bibinfo{number}{2} (\bibinfo{year}{1991}),
  \bibinfo{pages}{172--178}.
\newblock
\showISSN{00220485, 21524068}
\urldef\tempurl%
\url{http://www.jstor.org/stable/1182422}
\showURL{%
\tempurl}


\bibitem[\protect\citeauthoryear{{M}c{F}ee, {R}affel, {L}iang, {E}llis,
  {M}c{V}icar, {B}attenberg, and {N}ieto}{{M}c{F}ee et~al\mbox{.}}{2015}]%
        {McFee2015}
\bibfield{author}{\bibinfo{person}{{B}. {M}c{F}ee}, \bibinfo{person}{{C}.
  {R}affel}, \bibinfo{person}{{D}. {L}iang}, \bibinfo{person}{{D}.~{P}.{W}.
  {E}llis}, \bibinfo{person}{{M}. {M}c{V}icar}, \bibinfo{person}{{E}.
  {B}attenberg}, {and} \bibinfo{person}{{O}. {N}ieto}.}
  \bibinfo{year}{2015}\natexlab{}.
\newblock \showarticletitle{librosa: {A}udio and {M}usic {S}ignal {A}nalysis in
  {P}ython}. In \bibinfo{booktitle}{\emph{{P}roceedings of the 14th {P}ython in
  {S}cience {C}onference}}, \bibfield{editor}{\bibinfo{person}{{K}. {H}uff}
  {and} \bibinfo{person}{{J}. {B}ergstra}} (Eds.). \bibinfo{pages}{18 -- 25}.
\newblock


\bibitem[\protect\citeauthoryear{Nair and Hinton}{Nair and Hinton}{2010}]%
        {ReLU}
\bibfield{author}{\bibinfo{person}{Vinod Nair} {and}
  \bibinfo{person}{Geoffrey~E. Hinton}.} \bibinfo{year}{2010}\natexlab{}.
\newblock \showarticletitle{Rectified Linear Units Improve Restricted Boltzmann
  Machines}. In \bibinfo{booktitle}{\emph{Proceedings of the 27th International
  Conference on Machine Learning}} \emph{(\bibinfo{series}{ICML'10})}.
  \bibinfo{publisher}{Omnipress}, \bibinfo{address}{USA},
  \bibinfo{pages}{807--814}.
\newblock
\showISBNx{978-1-60558-907-7}
\urldef\tempurl%
\url{http://dl.acm.org/citation.cfm?id=3104322.3104425}
\showURL{%
\tempurl}


\bibitem[\protect\citeauthoryear{Shafique, Theocharides, Bouganis, Hanif,
  Khalid, Haf{\i}z, and Rehman}{Shafique et~al\mbox{.}}{2018}]%
        {shafique2018overview}
\bibfield{author}{\bibinfo{person}{Muhammad Shafique},
  \bibinfo{person}{Theocharis Theocharides}, \bibinfo{person}{Christos-Savvas
  Bouganis}, \bibinfo{person}{Muhammad~Abdullah Hanif}, \bibinfo{person}{Faiq
  Khalid}, \bibinfo{person}{Rehan Haf{\i}z}, {and} \bibinfo{person}{Semeen
  Rehman}.} \bibinfo{year}{2018}\natexlab{}.
\newblock \showarticletitle{An overview of next-generation architectures for
  machine learning: Roadmap, opportunities and challenges in the IoT era}. In
  \bibinfo{booktitle}{\emph{2018 Design, Automation \& Test in Europe
  Conference \& Exhibition (DATE)}}. IEEE, \bibinfo{pages}{827--832}.
\newblock


\bibitem[\protect\citeauthoryear{Tzanetakis and Cook}{Tzanetakis and
  Cook}{2002}]%
        {1021072}
\bibfield{author}{\bibinfo{person}{G. Tzanetakis} {and} \bibinfo{person}{P.
  Cook}.} \bibinfo{year}{2002}\natexlab{}.
\newblock \showarticletitle{Musical genre classification of audio signals}.
\newblock \bibinfo{journal}{\emph{IEEE Transactions on Speech and Audio
  Processing}} \bibinfo{volume}{10}, \bibinfo{number}{5} (\bibinfo{date}{July}
  \bibinfo{year}{2002}), \bibinfo{pages}{293--302}.
\newblock
\showISSN{1063-6676}
\urldef\tempurl%
\url{https://doi.org/10.1109/TSA.2002.800560}
\showDOI{\tempurl}


\bibitem[\protect\citeauthoryear{{Yao}, {Hao}, {Zhao}, {Piao}, {Shao}, {Liu},
  {Liu}, {Hu}, {Weerakoon}, {Jayarajah}, {Misra}, and {Abdelzaher}}{{Yao}
  et~al\mbox{.}}{2019}]%
        {Eugene}
\bibfield{author}{\bibinfo{person}{S. {Yao}}, \bibinfo{person}{Y. {Hao}},
  \bibinfo{person}{Y. {Zhao}}, \bibinfo{person}{A. {Piao}}, \bibinfo{person}{H.
  {Shao}}, \bibinfo{person}{D. {Liu}}, \bibinfo{person}{S. {Liu}},
  \bibinfo{person}{S. {Hu}}, \bibinfo{person}{D. {Weerakoon}},
  \bibinfo{person}{K. {Jayarajah}}, \bibinfo{person}{A. {Misra}}, {and}
  \bibinfo{person}{T. {Abdelzaher}}.} \bibinfo{year}{2019}\natexlab{}.
\newblock \showarticletitle{Eugene: Towards Deep Intelligence as a Service}. In
  \bibinfo{booktitle}{\emph{2019 IEEE 39th International Conference on
  Distributed Computing Systems (ICDCS)}}. \bibinfo{pages}{1630--1640}.
\newblock
\urldef\tempurl%
\url{https://doi.org/10.1109/ICDCS.2019.00162}
\showDOI{\tempurl}


\bibitem[\protect\citeauthoryear{{Zhang}, {Suda}, {Lai}, and {Chandra}}{{Zhang}
  et~al\mbox{.}}{2018}]%
        {Zhang2018}
\bibfield{author}{\bibinfo{person}{Y. {Zhang}}, \bibinfo{person}{N. {Suda}},
  \bibinfo{person}{L. {Lai}}, {and} \bibinfo{person}{V. {Chandra}}.}
  \bibinfo{year}{2018}\natexlab{}.
\newblock \showarticletitle{{Hello Edge: Keyword Spotting on
  Microcontrollers}}.
\newblock \bibinfo{journal}{\emph{ArXiv e-prints}} (\bibinfo{date}{Feb.}
  \bibinfo{year}{2018}).
\newblock
\showeprint[arxiv]{cs.SD/1711.07128}


\end{thebibliography}

\appendix

\section{Neural Network Architectures}
\begin{table*}[h]
\centering
\caption{Pareto optimal CNN architectures for KWS~\cite{nas_hpcs2019}}
\begin{tabular}{ rccccccccc }
Model & conv1 & conv2 & conv3 & conv4 & conv5 & conv6 & TOP-1 & M$FP_{ops}$  & Size (KB) \\
\toprule
seed & 4x10, 100 & 3x3, 100 & 3x3, 100 & 3x3, 100 & 3x3, 100 & 3x3, 100 & 94.2\% & 581.1 & 1832 \\
\midrule
kws1 & 3x3, 40 & 3x3, 30 & 1x1, 30 & 5x5, 50 & 5x5, 50 & 5x5, 50 & 95.1\% & 223.4 & 707.0\\
kws3 & 5x5, 50 & 1x1, 30 & 5x5, 40 & 3x3, 20 & 5x5, 30 & 3x3, 50 & 94.1\% & 87.6  & 282.1\\
kws9 & 5x5, 50 & 1x1, 20 & 1x1, 50 & 3x3, 20 & 5x5, 20 & 3x3, 40 & 93.4\% & 37.7 & 125.3\\
\bottomrule
\end{tabular}
\label{tab:pareto_best_CNN}
\end{table*}

\begin{table*}[h]
\centering
\caption{Optimized DS\_CNN architectures based on CNN models}
\begin{tabular}{ rccccccccc }
Model & conv1 & conv2 & conv3 & conv4 & conv5 & conv6 & TOP-1 & M$FP_{ops}$ & Size (KB) \\
\toprule
seed & 4x10, 100 & 3x3, 100 & 3x3, 100 & 3x3, 100 & 3x3, 100 & 3x3, 100 & 90.6\% & 69.9 & 1017 \\
\midrule
ds\_kws1 & 3x3, 40 & 3x3, 30 & 1x1, 30 & 5x5, 50 & 5x5, 50 & 5x5, 50 &  92.6\% & 11.9 & 61.5\\
ds\_kws3 & 5x5, 50 & 1x1, 30 & 5x5, 40 & 3x3, 20 & 5x5, 30 & 3x3, 50 &  91.2\% & 9.7  & 48.4\\
ds\_kws9 & 5x5, 50 & 1x1, 20 & 1x1, 50 & 3x3, 20 & 5x5, 20 & 3x3, 40 &  91.3\% & 7.0 & 39.0\\
\bottomrule
\end{tabular}
\label{tab:pareto_best_DSCNN}
\end{table*}

\end{document}